\documentclass{article}

\usepackage{microtype}
\usepackage{graphicx}
\usepackage{subfigure}
\usepackage{booktabs} 

\usepackage{lineno}

\usepackage[accepted]{icml2024}

\usepackage{hyperref}

\usepackage{multirow}
\usepackage{colortbl}

\usepackage{amsmath}
\usepackage{amssymb}
\usepackage{mathtools}
\usepackage{amsthm}

\usepackage[capitalize,noabbrev]{cleveref}

\theoremstyle{plain}

\theoremstyle{definition}

\theoremstyle{remark}

\usepackage[textsize=tiny]{todonotes}

\icmltitlerunning{Towards Few-shot Out-of-Distribution Detection}

\begin{document}

\twocolumn[
\icmltitle{Towards Few-shot Out-of-Distribution Detection}



\icmlsetsymbol{equal}{*}

\begin{icmlauthorlist}
\icmlauthor{Jiuqing Dong}{1}
\icmlauthor{Yongbin Gao}{2}
\icmlauthor{Heng Zhou}{1}
\icmlauthor{Jun Cen}{3}
\icmlauthor{Yifan Yao}{1}
\icmlauthor{Sook Yoon}{4}
\icmlauthor{Dong Sun Park}{1}
\end{icmlauthorlist}

\icmlaffiliation{1}{Jeonbuk National University, Jeonju, South Korea}
\icmlaffiliation{2}{Shanghai University of Engineering Science, Shanghai, China}
\icmlaffiliation{3}{The Hong Kong University of Science and Technology, Hong Kong, China}
\icmlaffiliation{4}{Mokpo National University, Mokpo, South Korea}

\icmlcorrespondingauthor{Sook Yoon}{syoon@mokpu.ac.kr}
\icmlcorrespondingauthor{Dong Sun Park}{dspark@jbnu.ac.kr}

\icmlkeywords{Machine Learning, ICML}

\vskip 0.3in
]
 



\begin{abstract}
Despite the notable advancements in current OOD detection methodologies, we find they are far from reaching satisfactory performance under the scarcity of training samples.
In this context, we introduce a novel few-shot OOD detection benchmark, carefully constructed to address this gap. Our empirical analysis reveals the superiority of Parameter-Efficient Fine-Tuning (PEFT) strategies, such as visual prompt tuning and visual adapter tuning, over conventional techniques, including fully fine-tuning and linear probing tuning in the few-shot OOD detection task. Recognizing some crucial information from the pre-trained model, which is pivotal for OOD detection, may be lost during the fine-tuning process, we propose a method termed Domain-Specific and General Knowledge Fusion (DSGF). This method is the
first time to strengthen fine-tuned features with original pre-trained features to recover the general information lost during fine-tuning for better OOD detection. Our experiments show that the integration of DSGF significantly enhances the few-shot OOD detection capabilities across various methods and fine-tuning methodologies, including fully fine-tuning, visual adapter tuning, and visual prompt tuning. The code will be released.
\vspace{-0.6cm}
\end{abstract}

\section{Introduction}
\label{sec:intro}
A reliable visual recognition system should not only provide accurate predictions within known contexts but also be able to detect unknown instances and reject them~\cite{yang2022openood}. This capability, known as out-of-distribution (OOD) detection, is imperative for maintaining AI safety, as it prevents recognition systems from erroneously processing inputs alien to their training tasks~\cite{amodei2016concrete}. For instance, a well-trained species detector should correctly identify common species within an ecosystem and alert them to invasive species rather than blindly classifying them into existing species categories~\cite{cultrera2023leveraging}. Similarly, in autonomous driving scenarios, the system must detect and respond to unforeseen environmental conditions or novel objects, potentially by alerting the driver for intervention~\cite{henriksson2023out}.

\begin{figure}[t]
  \centering
        \includegraphics[width=0.9\linewidth]{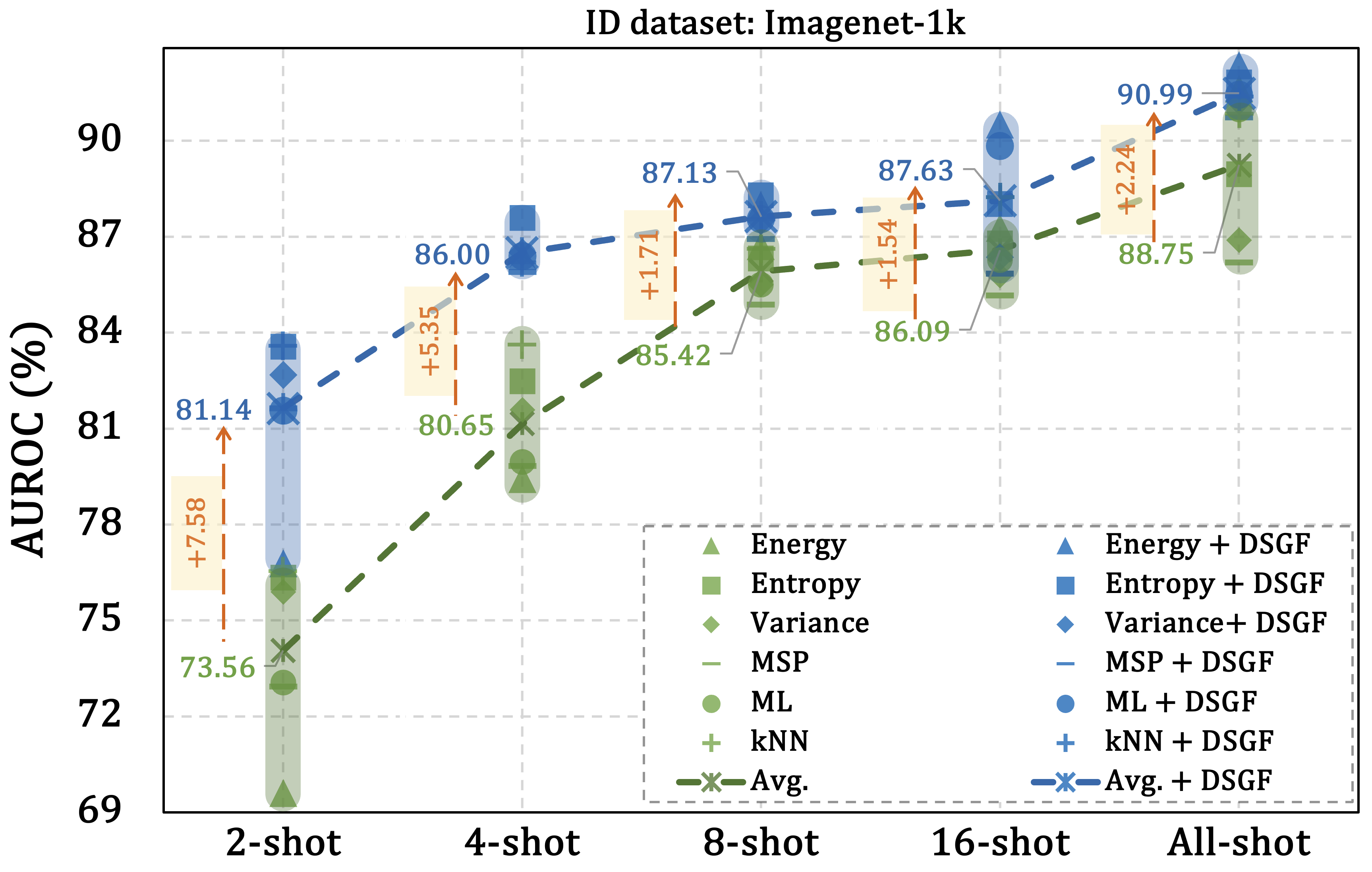}
   \caption{Comparison of different OOD detection methods in the FS-OOD detection task. Our DSGF significantly improves the performance of all baseline methods. `Avg.' represents the average of six OOD detection baseline methods, and ` + DSGF' denotes deploying our method.\vspace{-0.6cm}}
   \label{fig1}
\end{figure}

Current approaches for the OOD detection task are primarily trained on the entire in-distribution (ID) dataset~\cite{hendrycks2016baseline, liang2018enhancing, ren2019likelihood, liu2020energy, lin2021mood, hendrycks2022scaling, sun2022out}. However, obtaining a vast amount of labeled data is sometimes impractical. For instance, gathering extensive samples for each species in a biodiverse ecosystem is formidable~\cite{lu2022few}. Therefore, how to keep the strong OOD detection ability with limited ID training samples is significant. To explore this problem, we first construct a comprehensive few-shot out-of-distribution (FS-OOD) detection benchmark in this paper.  Our findings illustrate that the OOD detection performance drops quickly with the decrease of training samples, as shown in Fig.~\ref{fig1}, meaning that current OOD methods are not robust enough with limited training samples and designing a method under the few-shot condition is essential.

In the context of few-shot condition, large-scale pre-trained base models have shown remarkable performance gains across various downstream tasks~\cite{radford2021learning, zhang2022glipv2, oquab2023dinov2, kirillov2023segment}. However, the optimal approach to fine-tune these models for enhanced OOD detection with limited training samples remains an under-explored area. We notice that Parameter-Efficient Fine-Tuning (PEFT) is a promising strategy for maximizing the utility of pre-trained models within the constraints of limited training data~\cite{fu2023effectiveness,zhou2022learning, zhou2022conditional}. Therefore, in addition to the traditional fully fine-tuning (FFT) and linear probing tuning (LPT)~\cite{kornblith2019better}, we also include two PEFT methods, \textit{i.e.,} visual prompt tuning (VPT)~\cite{jia2022visual} and visual adapter tuning (VAT)~\cite{liu2020transfer} in our FS-OOD detection benchmark.

The experiments show that PEFT methods exhibit a great advantage in the FS-OOD detection task, as shown in Table~\ref{Table1},~\ref{Table2},~\ref{Table3}. Therefore, we hypothesize that the general knowledge stored in the pre-trained model is significant for OOD detection. PEFT methods freeze the pre-trained weights and only fine-tune a small set of additional parameters, while FFT fine-tunes all parameters during training. Consequently, PEFT methods retain the general knowledge inherent in the pre-trained model to a greater extent, thereby potentially enhancing OOD detection capabilities. In contrast, FFT may inadvertently lead to losing this acquired general knowledge. This hypothesis is also testified by another experiment that non-parametric OOD detection methods like k-th nearest neighbor (k-NN)~\cite{sun2022out} could achieve better performance using the frozen pre-trained model than the fully fine-tuned model under the few-shot setting, as shown in Fig~\ref{Fig3}.  Based on this hypothesis, we propose the Domain-Specific and General knowledge Fusion (DSGF) method, which explicitly involves the general knowledge of the original pre-trained model to strengthen the OOD detection performance. Experiment results demonstrate that DSGF can significantly enhance the performance of existing OOD detection methods, as shown in Fig.~\ref{fig1}.

%
%

Our main contributions are threefold: \textbf{(i).} We are the first to establish a comprehensive benchmark of few-shot out-of-distribution detection. In addition to the traditional fully fine-tuning and linear probing tuning, we also incorporate PEFT methods, including visual adapter tuning and visual prompt tuning. \textbf{(ii).} We propose the Domain-Specific and General knowledge Fusion (DSGF) method, which is the first time to strengthen fine-tuned features with original pre-trained features to recover the general information potentially lost during downstream fine-tuning for OOD detection. \textbf{(iii).} Experiment results show that DSGF is a versatile and universally applicable method capable of significantly improving the performance of various OOD detection methods across different fine-tuning paradigms, particularly under the few-shot setting.

\section{Related Work}
\label{sec:Related Work}
\subsection{Out-of-distribution Detection}
In the latest review on OOD detection,~\cite{yang2022openood} pointed out that the advantages of post-hoc detection methods lie in their ease of use without requiring modifications to the training procedure and objectives. One of the early works is the maximum softmax probability (MSP)~\cite{hendrycks2016baseline}, a post-hoc method that assumes ID samples typically exhibit higher softmax probabilities than OOD samples.~\cite{hendrycks2016baseline} also estimate the uncertainty score by calculating information entropy from the softmax probability distribution. Another study~\cite{hendrycks2022scaling} argues that methods relying on softmax confidence scores tend to be overconfident in the posterior distribution for OOD data. To address this issue, ~\cite{hendrycks2022scaling} advocates utilizing maximum logits to achieve OOD detection. Unlike~\cite{liang2018enhancing, hendrycks2022scaling},~\cite{liu2020energy} found that the energy score is theoretically consistent with the input's probability density and is less susceptible to the problem of overconfidence. Lin et al.~\cite{lin2021mood} provided a theoretical explanation from the perspective of likelihood, where test samples with lower energy scores are considered ID data, and vice versa. We reimplement these post-hoc methods in our FS-OOD detection benchmark. Additionally, we investigate the nonparametric method k-th nearest neighbor (k-NN)~\cite{sun2022out}, and find it can yield equal or superior performance using pre-trained model compared to using fully fine-tuned model under few-shot scenarios.

\begin{figure*}[t]
\vspace{-0.2cm}
  \centering
    \includegraphics[width=0.88\linewidth]{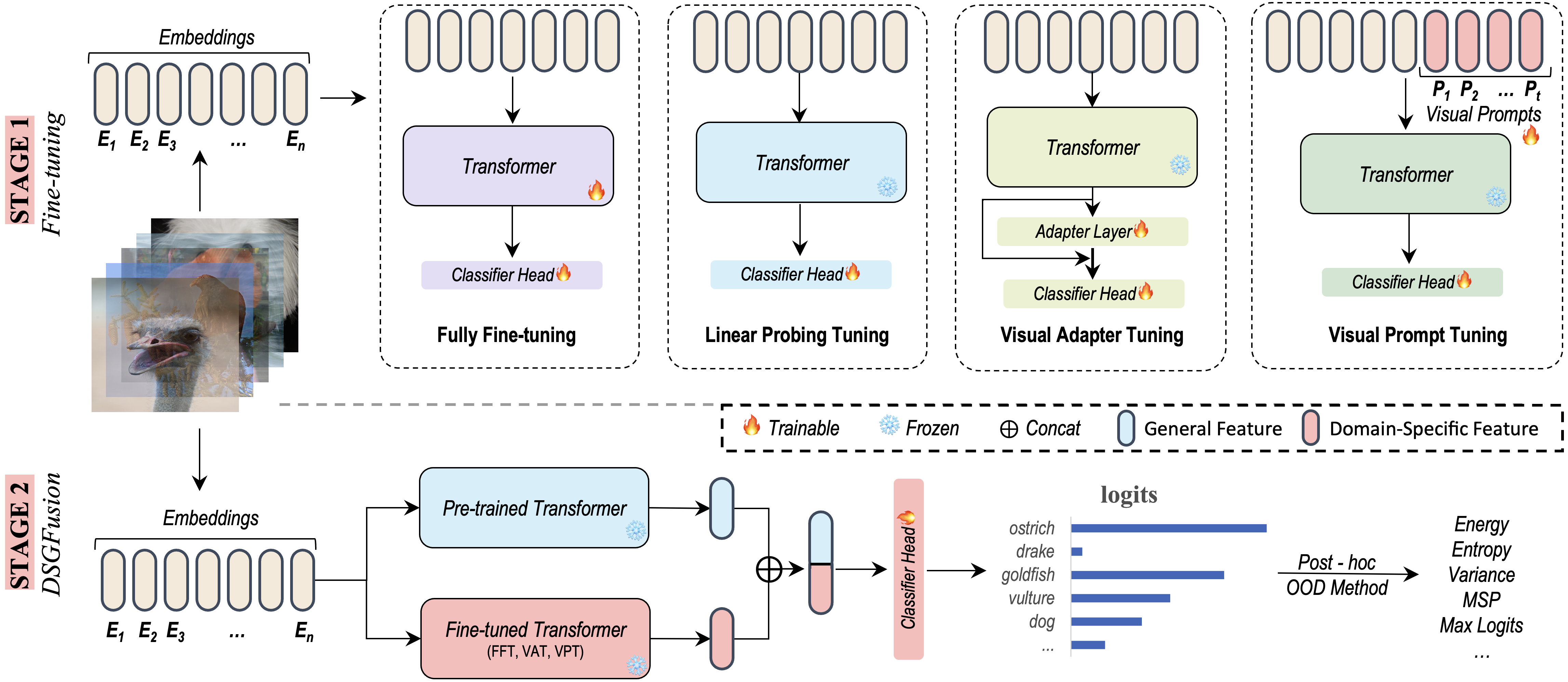}\vspace{-0.2cm}
   \caption{We include FFT, LPT, VAT, and VPT in our few-shot OOD detection benchmark (Stage 1). Our DSFG method fuses the general feature of the pre-trained model and the domain-specific feature of the fine-tuned model for better OOD detection performance (Stage 2).\vspace{-0.6cm}}
   \label{Fig2}
\end{figure*}

Recently, a series of works have explored OOD detection based on CLIP backbone~\cite{radford2021learning, ming2022delving, miyai2023locoop}. Compared to them, our FS-OOD uses purely vision backbone and does not require the language model, which is more
convenient and lightweight solution suitable for some real-world scenarios.


\subsection{Parameter Efficient Fine-tuning}
A series of large-scale pre-trained models have been released to enhance the performance of various downstream tasks~\cite{jiang2023rethinking}. The escalating scale of these models has precipitated a surge in the computational costs associated with their fine-tuning, manifested in the form of increased GPU memory requirements and extended training time. Parameter-Efficient Fine-Tuning (PEFT) has emerged as a viable alternative in response to these challenges, demonstrating competitive results relative to conventional fully fine-tuning~\cite{fu2023effectiveness}, particularly in few-shot learning~\cite{zhou2022learning, zhou2022conditional}. Consequently, in addition to traditional fine-tuning methods FFT and LPT, we also investigate two classic PEFT methods in our FS-OOD detection benchmark: visual adapter tuning (VAT) and visual prompt tuning (VPT).

VAT [24] is a fine-tuning method based on feed-forward networks (FFN). These adapters are adept at tailoring pre-trained spatial features for specific domain applications. Besides, a visual adapter can also be inserted at appropriate positions to introduce temporal information in video tasks~\cite{yang2023aim}. Besides, prompt tuning~\cite{lester2021power} originates from natural language processing, which leverages trainable tokens (prompts) to enhance performance in downstream tasks. Jia et al.~\cite{jia2022visual}  extended this concept to vision tasks. VPT initializes adjustable prompt tokens and adds them to the original image tokens at the first or hidden layers. We follow the default parameters outlined in~\cite{jia2022visual} and insert ten trainable prompt tokens before each transformer block.

\section{FS-OOD Detection Benchmark}
\label{sec:benchmark}

\subsection{Problem Definition}
In this section, we define the OOD detection problem under the few-shot condition. The training set is $D^{train}=\{x_i, y_i\}_{i=1}^{M*N}$, where $x$, $y$, $M$, and $N$ denote the sample, label, number of training samples of each ID class, and the number of classes. $M \le 16$ in our experiments, so the training set is extremely small and thus called few-shot. We define the set of ID classes as $K = \{1,2,3,...,N\}$, so $y \in K$. 
We assume that there exists a set of OOD classes $U = \{N+1, ... \}$, which the model does not witness during training but may encounter during inference, and $K \cap U = \emptyset$.
We can define OOD detection as a binary classification problem which is formalized as follows:
\begin{equation}
    G_{\lambda}(x_i) = 
    \begin{cases}
    OOD & S(x_i) > \lambda \\
    ID  & S(x_i) \leq \lambda
    \end{cases} ,
  \label{eq1}
\end{equation}
where a higher score $S(x_i)$ for a sample $x_i$ indicates higher uncertainty. A sample with a score greater than the threshold $\lambda$ will be classified as OOD, and vice versa. In addition, correctly classifying the ID samples is also required.

\subsection{Dataset Configuration}
\textbf{Setting 1:} We utilize three datasets as ID data sources: Imagenet-1k~\cite{deng2009imagenet}, FOOD-101~\cite{bossard2014food}, and Oxford-PETS~\cite{parkhi2012cats}.
To construct the OOD dataset, \cite{huang2021mos} collected diverse subsets from SUN~\cite{xiao2010sun}, iNaturalist~\cite{van2018inaturalist}, Places~\cite{zhou2017places}, and Texture~\cite{cimpoi2014describing} as a large-scale OOD dataset for Imagenet-1k. These datasets are carefully curated, with non-overlapping categories in their test sets compared to Imagenet-1k.

\noindent \textbf{Setting 2:} 
We use CIFAR-100~\cite{krizhevsky2009learning} as the ID dataset and Tiny-Imagenet-200~\cite{le2015tiny} as the OOD dataset, constituting a more challenging evaluation setup. In this setting, OOD datasets have semantic shifts compared with ID datasets, while in Setting 1 OOD, datasets mainly have obvious covariate (domain) shifts. 

In the few-shot setting, the number of training samples of each ID class is set to 2, 4, 8, and 16. Besides, we provide a brief overview of each dataset in Appendix~\ref{AppendixA.1}.


\begin{table*}[t]
\vspace{-0.5cm}
  \centering
  \caption{The result on the Imagenet-1k dataset. We compute the average FPR@95 and AUROC of four OOD datasets.}
  \label{Table1}
    \fontsize{7}{8}\selectfont
  \begin{tabular}{cccccccc}
    \hline
    \multirow{2}{*}{Shot} & \multirow{2}{*}{Method}
    & \multicolumn{6}{c}{FPR@95 $\downarrow$ / AUROC $\uparrow$ $(\%)$} \\
    \cline{3-8}
    & {} & Energy & Entropy & Variance & MSP	& Max-Logits & k-NN	 \\
    \hline
    \multirow{4}{*}{2-shot}    
        & Fully Fine-tuning           & 88.97 / 69.11 & 75.39 / 75.83 & 74.60 / 75.40 & 81.71 / 72.44 & 81.57 / 72.56 & 75.84 / 76.04  \\
        & Linear Probing Tuning       & 50.09 / 87.59 & 46.05 / 87.82 & 46.92 / 87.37 & 50.33 / 86.76 & 48.90 / 87.44 & 55.58 / 85.89  \\
        & Visual Adapter Tuning       & 61.94 / 85.39 & 46.61 / 87.08 & 50.80 / 86.06 & 54.39 / 85.33 & 52.99 / 86.17 & 53.02 / 85.73  \\
        & Visual Prompt Tuning        & 46.92 / 87.06 & 45.34 / 87.50 & 45.68 / 87.34 & 49.59 / 86.72 & 48.86 / 86.93 & 54.25 / 86.09  \\
    \midrule
    \multirow{4}{*}{4-shot}    
        & Fully Fine-tuning            & 73.32 / 78.93 & 58.54 / 81.97 & 61.07 / 81.08 & 69.64 / 79.33 & 69.21 / 79.46 & 61.88 / 83.12  \\
        & Linear Probing Tuning        & 39.70 / 89.86 & 39.20 / 89.67 & 42.59 / 91.80 & 45.44 / 88.54 & 42.67 / 89.44 & 55.11 / 87.19  \\
        & Visual Adapter Tuning        & 42.95 / 89.51 & 38.61 / 90.24 & 41.89 / 89.59 & 45.00 / 89.05 & 43.47 / 89.63 & 49.20 / 88.24  \\
        & Visual Prompt Tuning         & 48.90 / 87.36 & 47.60 / 88.19 & 48.18 / 88.43 & 50.20 / 88.18 & 49.49 / 88.23 & 53.42 / 87.99  \\
    \midrule
    \multirow{4}{*}{8-shot}    
        & Fully Fine-tuning           & 48.57 / 86.21 & 49.53 / 85.83 & 52.84 / 85.01 & 56.25 / 84.37 & 55.31 / 85.00 & 54.11 / 86.12  \\
        & Linear Probing Tuning       & 37.38 / 90.89 & 38.27 / 90.46 & 41.45 / 89.75 & 44.19 / 89.21 & 41.94 / 90.10 & 56.89 / 86.68  \\
        & Visual Adapter Tuning       & 38.39 / 90.62 & 38.40 / 90.63 & 41.94 / 89.95 & 44.20 / 89.43 & 42.35 / 90.25 & 49.35 / 88.43  \\
        & Visual Prompt Tuning        & 49.20 / 87.75 & 48.64 / 88.22 & 49.84 / 88.15 & 51.60 / 87.75 & 50.69 / 88.03 & 58.39 / 86.39  \\
    \midrule
    \multirow{4}{*}{16-shot}    
        & Fully Fine-tuning           & 48.00 / 86.77 & 49.20 / 86.30 & 51.77 / 85.29 & 53.36 / 84.66 & 52.48 / 85.78 & 49.81 / 87.73  \\        
        & Linear Probing Tuning       & 34.62 / 91.22 & 37.11 / 90.49 & 40.77 / 89.62 & 43.07 / 89.04 & 40.39 / 92.71 & 57.67 / 86.60  \\
        & Visual Adapter Tuning       & 34.85 / 91.55 & 37.22 / 89.81 & 40.59 / 89.81 & 42.44 / 89.28 & 39.47 / 90.92 & 47.30 / 89.15  \\
        & Visual Prompt Tuning        & 40.90 / 89.86 & 41.44 / 90.06 & 43.54 / 89.78 & 45.55 / 89.34 & 44.38 / 89.78 & 51.32 / 88.45  \\
    \midrule
    \multirow{4}{*}{All-shot}    
        & Fully Fine-tuning           & 39.09 / 91.13 & 43.67 / 86.39 & 49.02 / 86.39 & 51.74 / 85.69 & 42.61 / 90.48 & 43.28 / 90.35  \\
        & Linear Probing Tuning       & 34.38 / 91.71 & 34.10 / 90.90 & 37.02 / 90.90 & 41.49 / 90.22 & 40.45 / 90.50 & 62.58 / 85.39  \\
        & Visual Adapter Tuning       & 41.01 / 90.97 & 38.72 / 88.69 & 43.43 / 88.69 & 46.21 / 88.03 & 45.23 / 90.28 & 59.58 / 86.08  \\
        & Visual Prompt Tuning        & 39.86 / 90.21 & 41.89 / 89.60 & 43.16 / 89.60 & 44.59 / 89.08 & 43.21 / 89.82 & 64.52 / 87.00  \\
    \hline
  \end{tabular}
\vspace{-0.3cm}
\end{table*}

\subsection{Benchmark}

To effectively leverage large-scale pre-trained models to address FS-OOD detection tasks, this paper investigates the performance of different OOD detection methods under four fine-tuning paradigms and establishes a comprehensive benchmark. The details are as follows.

\textbf{Fine-tuning Paradigms.}
\textit{Fully Fine-Tuning (FFT)} updates all backbone and classification head parameters. While FFT allows for comprehensive adaptation to the target domain, it typically demands more data and computational resources. 

\textit{Linear Probing Tuning (LPT)} freezes the backbone and only updates classification head parameters. LPT is more suitable than FFT for few-shot learning, as a small number of samples have a distribution bias problem and cannot provide representative information for one class. In such cases, FFT using these small groups of training samples leads to overfitting, while LPT retains the general feature of the pre-trained model to alleviate this problem. 

\textit{Visual Adapter Tuning (VAT)} inserts lightweight adapter modules into each Transformer layer to facilitate task-specific learning. By freezing the pre-trained model's backbone, visual adapter tuning can preserve the general feature extraction capabilities of the pre-trained model to a significant extent. In this paper, we insert a basic multilayer perceptron (MLP) module as a feed-forward network~\cite{houlsby2019parameter, pfeiffer2020adapterhub} with residual connection inside Transformer layers.

\textit{Visual Prompt Tuning (VPT)} introduces only a small number of trainable tokens (less than $1\%$ of model parameters) into the input space to adapt the model to the current task. It also preserves the general knowledge of the pre-trained model by keeping the backbone frozen. In this paper, we insert ten learnable prompt tokens inside each Transformer layer and follow default settings in~\cite{jia2022visual}.

\noindent \textbf{OOD Detection Baselines.}
The classification head can be considered as a feature mapping that maps the input image's feature to the logits values. Post-hoc methods for OOD detection estimate the uncertainty scores of ID and OOD data by processing these logits values. We reproduce several classical OOD detection methods: energy~\cite{liu2020energy}, entropy~\cite{hendrycks2016baseline}, variance~\cite{ryu2018out}, maximum softmax probability~\cite{hendrycks2016baseline}, and max-logits~\cite{hendrycks2022scaling}. We also implement the k-NN method~\cite{sun2022out}, which is a non-parametric method that permits training-free by directly calculating the feature similarity between the test samples and the training dataset. The formal formulas of these methods to calculate the uncertainty scores are shown in Appendix~\ref{AppendixA.2}.

\noindent \textbf{Evaluation Metrics.}
We employ \textit{FPR@95}~\cite{du2021vos} and \textit{AUROC}~\cite{powers2011evaluation} as evaluation metrics for OOD detection. Besides, we use \textit{In-Distribution Accuracy (ID Acc.)}~\cite{gunawardana2009survey} as the metric to evaluate the accuracy of the ID dataset. More details are elaborated in Appendix~\ref{AppendixA.3}.

\begin{table*}[t]
\vspace{-0.5cm}
  \centering
  \caption{The result on the FOOD-101 dataset. We compute the average FPR@95 and AUROC of four OOD datasets.}
  \label{Table2}
    \fontsize{7}{8}\selectfont
  \begin{tabular}{cccccccc}
    \hline
    \multirow{2}{*}{Shot} & \multirow{2}{*}{Method}
    & \multicolumn{6}{c}{FPR@95 $\downarrow$ / AUROC $\uparrow$ $(\%)$} \\
    \cline{3-8}
    & {} & Energy & Entropy & Variance & MSP	& Max-Logits & k-NN	 \\
    \hline
    \multirow{4}{*}{2-shot}    
        & Fully Fine-tuning           & 17.40 / 95.24 & 23.72 / 93.05 & 32.31 / 90.42 & 45.54 / 87.29 & 31.38 / 91.68 & 15.00 / 95.98 \\
        & Linear Probing Tuning       &  1.33 / 99.60 &  5.59 / 98.53 &  9.84 / 97.52 & 14.52 / 96.56 &  2.52 / 99.37 &  1.07 / 99.72 \\
        & Visual Adapter Tuning       &  3.26 / 99.00 &  2.67 / 99.32 &  6.00 / 98.71 & 12.44 / 97.58 &  6.37 / 98.70 &  2.10 / 99.51 \\
        & Visual Prompt Tuning        &  2.43 / 99.37 &  4.30 / 98.90 &  6.69 / 98.27 & 14.96 / 96.64 & 10.16 / 97.69 &  5.70 / 98.71 \\
    \midrule
    \multirow{4}{*}{4-shot}           
        & Fully Fine-tuning           & 9.34 / 97.70  &	14.96 /	96.11 &	21.51 /	94.36 &	31.41 /	92.27 &	18.05 /	95.75 &	8.92 / 97.94  \\
        & Linear Probing Tuning       &  0.91 / 99.63 &  2.39 / 99.21 &  4.73 / 98.69 &  8.61 / 97.95 &  1.67 / 99.43 &  1.39 / 99.57 \\
        & Visual Adapter Tuning       &  3.33 / 98.92 &  2.08 / 99.37 &  4.35 / 98.91 &  8.56 / 98.11 &  4.48 / 98.91 &  1.57 / 99.49 \\
        & Visual Prompt Tuning        &  1.11 / 99.65 &  1.76 / 99.51 &  3.04 / 99.26 &  6.66 / 98.56 &  3.72 / 99.11 &  2.79 / 99.31 \\
    \midrule
    \multirow{4}{*}{8-shot}    
        & Fully Fine-tuning           & 11.65 / 97.71 & 10.43 / 97.70 & 16.18 / 96.47 & 28.73 / 94.23 & 19.03 / 96.23 &  5.84 / 98.78 \\
        & Linear Probing Tuning       &  0.98 / 99.62 &  6.66 / 98.08 & 12.22 / 96.85 & 17.70 / 95.87 &  1.16 / 99.55 &  1.04 / 99.72 \\
        & Visual Adapter Tuning       &  1.01 / 99.60 &  1.45 / 99.51 &  2.87 / 99.22 &  5.33 / 98.76 &  1.49 / 99.48 &  1.25 / 99.70 \\
        & Visual Prompt Tuning        &  0.96 / 99.73 &  1.59 / 99.58 &  2.58 / 99.37 &  4.75 / 98.97 &  2.26 / 99.44 &  1.82 / 99.57 \\
    \midrule
    \multirow{4}{*}{16-shot}    
        & Fully Fine-tuning           &  2.47 / 99.32 &  4.83 / 98.90 &  7.93 / 98.33 & 13.42 / 97.43 &  6.25 / 98.66 &  2.34 / 99.36 \\        
        & Linear Probing Tuning       &  0.67 / 99.71 &  2.67 / 99.21 &  5.50 / 98.63 &  9.24 / 97.95 &  0.95 / 99.61 &  1.16 / 99.66 \\
        & Visual Adapter Tuning       &  0.87 / 99.68 &  1.52 / 99.52 &  3.03 / 99.23 &  5.16 / 98.83 &  1.06 / 99.59 &  1.18 / 99.64 \\
        & Visual Prompt Tuning        &  0.73 / 99.80 &  1.01 / 99.72 &  1.34 / 99.62 &  2.05 / 99.46 &  1.12 / 99.69 &  1.35 / 99.61 \\
    \midrule
    \multirow{4}{*}{All-shot}    
        & Fully Fine-tuning           &  3.46 / 99.20 & 10.92 / 98.09 & 17.20 / 97.50 & 18.76 / 97.17 &  3.65 / 99.11 &  3.56 / 99.15 \\
        & Linear Probing Tuning       &  0.78 / 99.61 &  1.55 / 99.58 &  3.10 / 99.08 &  3.39 / 98.57 &  1.22 / 99.43 &  2.65 / 99.33 \\
        & Visual Adapter Tuning       &  1.08 / 99.58 &  3.21 / 99.16 &  7.00 / 98.52 & 10.95 / 97.91 &  1.34 / 99.46 &  1.43 / 99.59 \\
        & Visual Prompt Tuning        &  2.08 / 99.58 &  2.89 / 99.36 &  3.78 / 99.16 &  4.72 / 99.00 &  2.60 / 99.46 &  3.80 / 99.16 \\
    \hline
  \end{tabular}
\vspace{-0.3cm}
\end{table*}

\noindent \textbf{Results.} We comprehensively evaluate the performance metrics, including the FPR@95 and the AUROC scores, for the Imagenet-1k, FOOD-101, and CIFAR-100 datasets. These results are detailed in Table~\ref{Table1},~\ref{Table2},~\ref{Table3}. Additionally, we report the ID Accuracy for these datasets in Table~\ref{Table4}. Please note that for Imagenet-1k and FOOD-101 datasets, we report the average FPR@95 and AUROC scores across four OOD datasets. The results on Oxford-PETS and the detailed results for each OOD dataset are provided in Appendix~\ref{AppendixC.2}. 

The main result is three-fold: \textbf{(i).} As the shot increases, the FPR@95 decreases, while AUROC and ID accuracy gradually increase. This trend is most pronounced in the fully fine-tuning setting. It indicates that keeping the high OOD detection performance with limited training samples is more challenging, illustrating the significance of our FS-OOD benchmark. \textbf{(ii).}  In the FS-OOD detection task, efficient fine-tuning methods significantly outperform fully fine-tuning across various metrics. For instance, Table~\ref{Table1} reveals that in the 2-shot experiments with the energy-based OOD detection method, LPT, VAT, and VPT significantly outperform FFT by 16\%-18\% on AUROC. Additionally, Table~\ref{Table4} further shows that LPT, VAT, and VPT achieve higher ID accuracy than FFT under few-shot settings. However, when the entire training dataset is available, FFT, VAT, and VPT exhibit comparable performance and notably outperform LPT. \textbf{(iii).} No single OOD detection method consistently outperforms the others in our benchmark. Rankings of OOD methods may vary significantly across different fine-tuning paradigms, different shot settings, and different datasets. It is worth noting that, except for the linear probing tuning, k-NN-based methods maintain relatively high AUROC scores in most settings. In LPT, the feature extractor does not update any parameters, making the comparison of k-NN scores with other OOD methods unfair.

\begin{table*}[t]
\vspace{-0.3cm}
  \centering
  \caption{The result on the CIFAR-100 dataset. Note that the OOD dataset is Tiny-Imagenet-200.}
  \label{Table3}
    \fontsize{7}{8}\selectfont
  \begin{tabular}{cccccccc}
    \hline
    \multirow{2}{*}{Shot} & \multirow{2}{*}{Method}
    & \multicolumn{6}{c}{FPR@95 $\downarrow$ / AUROC $\uparrow$ $(\%)$} \\
    \cline{3-8}
    & {} & Energy & Entropy & Variance & MSP	& Max-Logits & k-NN	 \\
    \hline
    \multirow{4}{*}{2-shot}    
        & Fully Fine-tuning           & 89.56 / 61.56 &	91.30 /	59.63 &	91.34 /	58.90 &	92.08 / 58.37 &	91.04 / 60.18 & 91.17 / 61.21   \\
        & Linear Probing Tuning       & 89.78 / 64.60 &	82.88 /	67.25 &	84.30 /	65.70 &	85.79 / 64.53 &	86.41 / 65.18 & 77.40 / 71.32   \\
        & Visual Adapter Tuning       & 79.48 / 72.54 &	80.89 /	71.21 &	82.29 /	69.79 &	84.06 / 67.93 &	82.59 / 69.52 & 84.03 / 69.49   \\
        & Visual Prompt Tuning        & 67.48 / 77.47 &	68.49 /	75.68 &	70.73 /	73.80 &	76.83 / 71.05 &	74.21 / 72.99 & 77.70 / 70.52   \\
    \midrule
    \multirow{4}{*}{4-shot}    
        & Fully Fine-tuning           & 85.93 / 67.34 &	84.95 /	66.19 &	86.28 /	65.21 &	87.79 / 63.93 &	86.08 / 65.60 & 82.44 / 68.50   \\
        & Linear Probing Tuning       & 90.47 / 67.07 &	78.47 /	71.23 &	80.74 /	69.56 &	82.85 / 68.26 &	85.08 / 68.53 & 74.63 / 74.22   \\
        & Visual Adapter Tuning       & 65.75 / 78.86 &	67.84 /	77.55 &	70.46 /	76.23 &	75.24 / 74.77 &	72.45 / 76.45 & 73.88 / 75.78   \\
        & Visual Prompt Tuning        & 65.92 / 78.76 &	66.47 /	76.51 &	68.94 /	74.28 &	75.17 / 71.73 &	71.63 / 74.46 & 76.84 / 71.34   \\
    \midrule
    \multirow{4}{*}{8-shot}    
        & Fully Fine-tuning           & 78.69 / 73.13 &	80.24 /	71.88 &	82.17 /	70.47 &	83.70 / 69.35 &	80.75 / 71.70 & 77.07 / 74.52   \\
        & Linear Probing Tuning       & 78.80 / 74.50 &	74.19 /	74.89 &	77.36 /	73.26 &	80.40 / 72.15 &	78.70 / 74.29 & 71.05 / 76.20   \\
        & Visual Adapter Tuning       & 65.10 / 80.84 &	67.96 /	79.67 &	69.41 /	78.59 &	72.00 / 77.73 &	69.21 / 79.55 & 67.24 / 79.41   \\
        & Visual Prompt Tuning        & 63.48 / 79.34 &	67.59 /	76.56 &	71.41 /	74.25 &	76.76 / 71.87 &	72.90 / 75.09 & 74.46 / 71.83   \\
    \midrule
    \multirow{4}{*}{16-shot}    
        & Fully Fine-tuning           & 72.80 / 76.84 &	73.37 /	75.87 &	75.43 /	74.71 &	78.07 / 73.89 &	75.74 / 75.88 & 73.99 / 76.69   \\
        & Linear Probing Tuning       & 68.14 / 79.64 &	69.21 /	77.35 &	72.41 /	75.62 &	74.81 / 74.73 &	69.47 / 78.99 & 67.21 / 78.54   \\
        & Visual Adapter Tuning       & 60.58 / 82.28 &	64.12 /	80.43 &	66.80 /	78.93 &	69.68 / 78.09 &	65.84 / 81.05 & 64.96 / 81.19   \\
        & Visual Prompt Tuning        & 62.47 / 79.62 &	66.74 /	76.00 &	70.98 /	73.40 &	76.45 / 71.35 &	71.11 / 75.30 & 74.44 / 72.07   \\
    \midrule
    \multirow{4}{*}{All-shot}    
        & Fully Fine-tuning           & 64.40 / 82.50 &	69.51 /	80.59 &	72.45 /	80.13 &	74.67 / 80.00 &	65.73 / 82.39 & 63.51 / 84.24   \\
        & Linear Probing Tuning       & 53.70 / 84.76 &	60.52 /	81.70 &	64.27 /	80.11 &	66.53 / 79.34 &	56.94 / 84.21 & 59.23 / 83.73   \\
        & Visual Adapter Tuning       & 53.93 / 84.92 &	59.37 /	82.27 &	63.07 /	80.59 &	64.84 / 79.85 &	58.06 / 83.99 & 55.07 / 85.20   \\
        & Visual Prompt Tuning        & 62.51 / 80.43 &	67.61 /	76.69 &	71.97 /	74.35 &	76.65 / 72.84 &	71.77 / 77.68 & 74.62 / 74.02   \\
    \hline
  \end{tabular}
\vspace{-0.4cm}
\end{table*}

\section{Method}
\label{sec:Methods}
\noindent{\bf General Knowledge for OOD detection.} We hypothesize that the reason that PEFT methods have clearly better OOD detection performance than FFT lies in the general knowledge of the pre-trained model. PEFT methods, including VAT and VPT, freeze the backbone during the fine-tuning process, which preserves the general knowledge of the original pre-trained model. In contrast, FFT inevitably loses some valuable general knowledge as it changes the parameters of the backbone. Zero-shot OOD detection method maximum concept matching (MCM)~\cite{ming2022delving} also shows that the pre-trained vision-language CLIP model~\cite{radford2021learning} achieves considerable OOD detection without fine-tuning,  which also supports our hypothesis that the general knowledge stored in the pre-trained model is significant for the OOD detection.

\begin{figure}[t]
  \centering
    \includegraphics[width=0.95\linewidth]{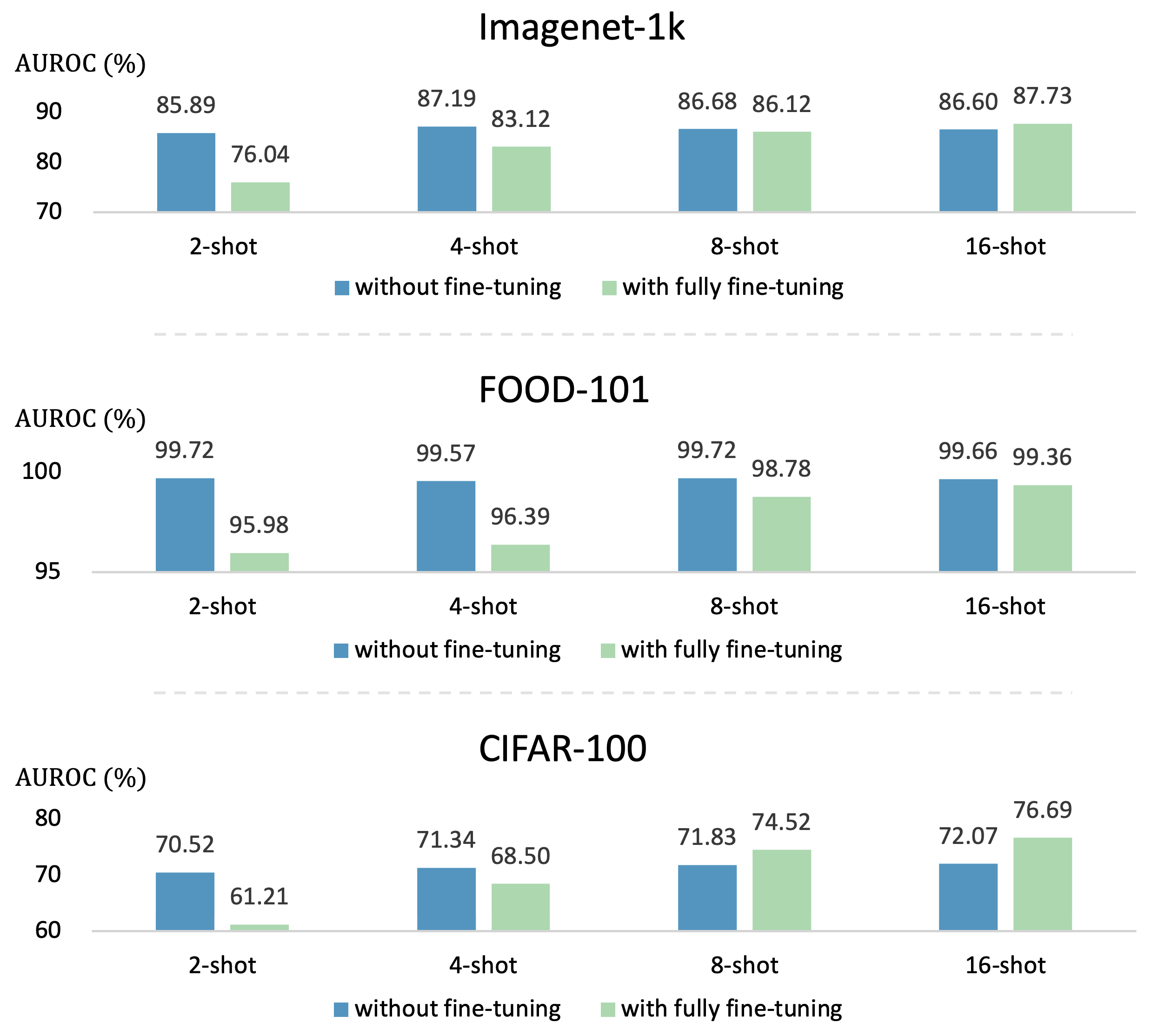}
   \caption{Comparison of AUROC scores between without and with fully fine-tuning for three ID datasets. OOD Scores are computed by the k-NN method.\vspace{-0.6cm}}
   \label{Fig3}
\end{figure}

Since MCM~\cite{ming2022delving} utilizes a pre-trained vision-language model for OOD detection without fine-tuning, we want to explore whether a purely pre-trained vision model could achieve OOD detection using only pre-trained general knowledge. We find that the non-parametric OOD detection method k-NN~\cite{sun2022out} could estimate the uncertainty score without fine-tuning. It calculates the feature distance between the testing and training samples for uncertainty estimation, and we directly utilize the output features of the original pre-trained model for this method. We observe that pre-trained models without fine-tuning have already achieved remarkable AUROC scores. As shown in Fig.~\ref{Fig3}, the AUROC of the original model without fine-tuning is approximately $86\%$, $99.7\%$, and $71\%$ when the ID dataset is ImageNet-1k, FOOD-101, and CIFAR-100. Furthermore, the untuned model even outperforms fully fine-tuned models in some few-shot settings. These indicate that general knowledge in the large-scale pre-trained models already achieves strong OOD detection capabilities.

\noindent{\bf Domain-specific Knowledge for OOD Detection.} The pre-trained model obtains the domain-specific knowledge during the fine-tuning process. The domain-specific knowledge is not only significant for downstream ID accuracy but also helpful for OOD detection. Table~\ref{Table1},~\ref{Table2},~\ref{Table3},~\ref{Table4} show that the OOD detection performance and ID accuracy keep increasing with more training samples for different baseline methods and fine-tuning paradigms.

\noindent{\bf Domain-Specific and General knowledge Fusion.} Both domain-specific and general knowledge are essential for OOD detection, but there is a trade-off between these two types of knowledge. Fine-tuning with more training samples could provide more domain-specific knowledge, but the model inevitably loses more general knowledge of the original pre-trained model. Although freezing the backbone, as PEFT methods do, is an effective strategy to keep the general knowledge, it could not involve all general knowledge in the output feature because of learnable adapters and prompts. So, we propose a simple yet effective method called Domain-Specific and General knowledge Fusion (DSGF) to solve this trade-off.

Stage 2 in Fig.~\ref{Fig2} illustrates our DSGF method. We aim to leverage the original pre-trained model's general knowledge and the fine-tuned model's domain-specific knowledge to address FS-OOD detection. Firstly, we obtain feature embeddings ${f}_{o}$ and ${f}_{{ft}}$ from original pre-trained model $M_o$ and fine-tuned model $M_{ft}$ for each image $X$:
\begin{equation}
  \label{eq2}
    {f}_{o} = M_o(X) \in \mathbb{R}^{d}\text{,} \quad {f}_{{ft}} = M_{ft}(X)\in \mathbb{R}^{d}\text{,}
\end{equation}
where $d$ denotes the dimension of the features. The model $M_{ft}$ is fine-tuned by FFT, VAT, or VPT. Then, we concatenate ${f}_{o}$ and ${f}_{{ft}}$ to obtain the fused feature ${f}_{{fs}}$:
\begin{equation}
  f_{fs} = \text{Concat}({f}_{o}, {f}_{{ft}})\in \mathbb{R}^{2*d}\text{,}
  \label{eq3}
\end{equation}
so that ${f}_{{fs}}$ contains general knowledge from the original pre-trained model $M_o$ and the domain-specific knowledge from the fine-tuned model $M_{ft}$. The logits $l_{fs}$ are produced by a fully-connected (FC) classifier:
\begin{equation}
  l_{fs} = \text{FC}(f_{fs})\in \mathbb{R}^{N}\text{.}
  \label{eq4}
\end{equation}
The loss $\mathcal{L}$ is defined as the cross-entropy loss:
\begin{equation}
  \mathcal{L} = - \text{log} \frac{\text{exp}(l_{fs}^y)}{\sum_{i=1}^{N}\text{exp}(l_{fs}^i)}\text{,}
  \label{eq5}
\end{equation}
where $y$ represents the ground-truth label.
During inference, we can deploy post-hoc OOD detection methods using logits $l_{fs}$ or k-NN method using ${f}_{{fs}}$.

\section{Experiments}
\label{sec:Experiment Result}

\noindent{\bf Implementation Details.} We use vision transformers (ViT)~\cite{dosovitskiy2020image} pre-trained on ImageNet-21k~\cite{deng2009imagenet} in this work. For STAGE 2 of DSGF, we train the new linear classification head for an extra 20 epochs. The optimal hyper-parameters typically do not share between datasets. We report the best hyper-parameters for each ID dataset in Appendix ~\ref{AppendixB}.


\noindent{\bf Results of DSGF.} We provide a detailed comparative analysis of the performance of our method against baselines, focusing on AUROC scores, as illustrated in Fig.~\ref{Fig4}. Moreover, a comparison of FPR@95 scores is visualized in Appendix~\ref{AppendixC.1}. To establish a complete FS-OOD detection benchmark, we report FPR@95 and AUROC scores for each pair of ID and OOD datasets in Appendix~\ref{AppendixC.2}.

\begin{figure*}[t]
\vspace{-0.2cm}
  \centering
    \includegraphics[width=0.92\linewidth]{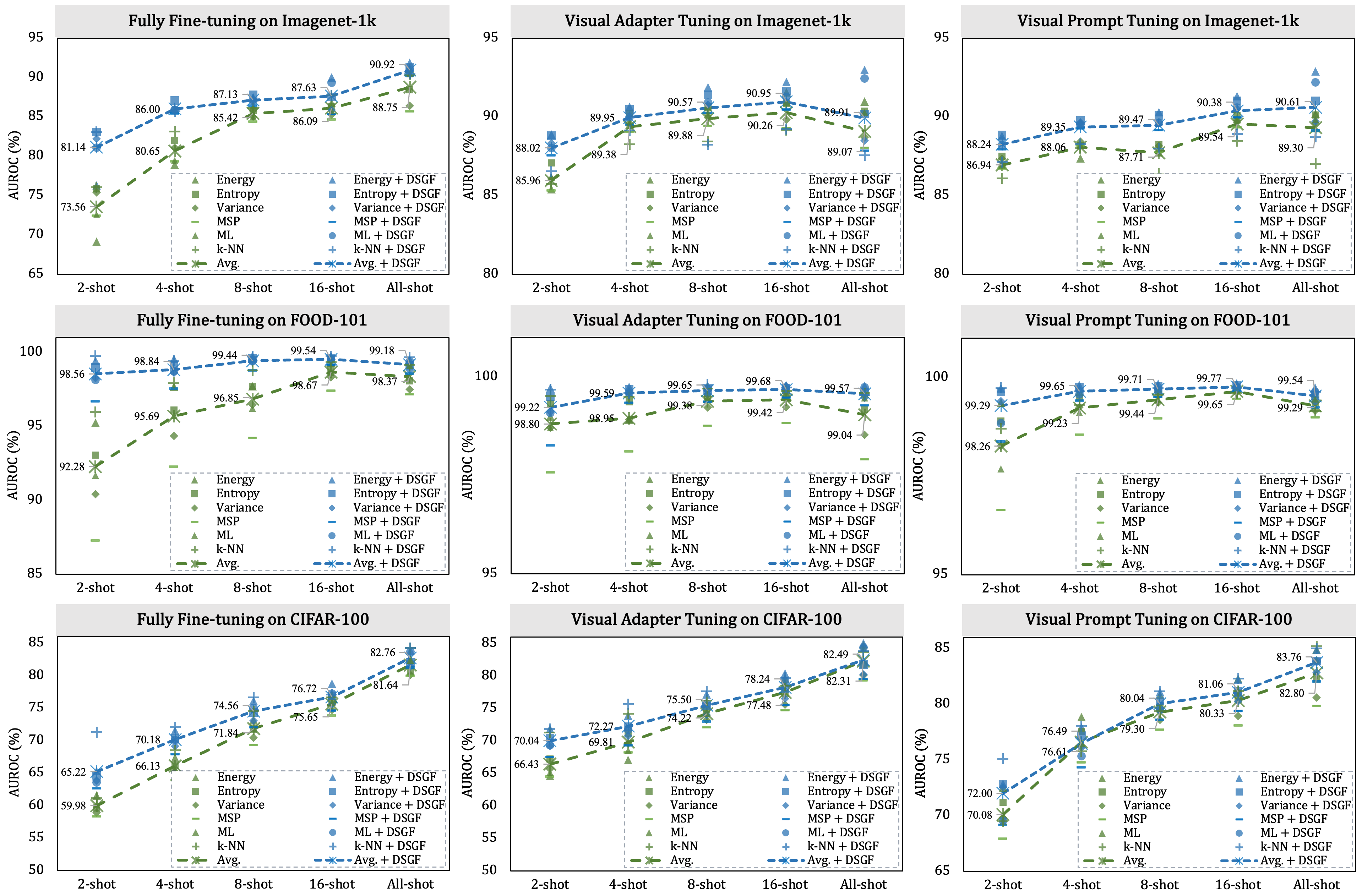}\vspace{-0.3cm}
  \caption{Main Results of FS-OOD detection on different tuning paradigms. Overall, our method achieves more performance gains in few-shot settings. `Avg.' represents the arithmetic average of six OOD score evaluation methods. `DSGF' denotes deploying our method.\vspace{-0.3cm}}
  \label{Fig4}
\end{figure*}

\begin{table}[t]
\vspace{-0.3cm}
  \centering
  \caption{Results of ID Accuracy on three datasets.}
  \label{Table4}
    \fontsize{7}{8}\selectfont
  \begin{tabular}{cccccc}
    \multirow{2}{*}{Method} & \multicolumn{5}{c}{ID Acc. ($\%$) (ID dataset: Imagenet-1k)} \\
    \cline{2-6}
                                        &2-shot     &4-shot     &8-shot     &16-shot    & All-shot  \\
    \hline
                            FFT	         & 54.91     & 63.30    & 65.26	    & 69.66	    & 80.1  \\
    \rowcolor{gray!25}      FFT + DSGF	 & 55.26     & 63.48    & 68.33	    & 71.61	    & 81.6   \\
    \hline   
                            VAT	         & 63.80     & 70.70    & 74.06	    & 75.91	    & 80.6   \\
    \rowcolor{gray!25}      VAT + DSGF	 & 64.91     & 70.75    & 73.59	    & 75.41     & 80.8  \\
    \hline   
                            VPT	         & 67.62     & 72.80    & 75.22	    & 76.28	    & 78.6   \\
    \rowcolor{gray!25}      VPT + DSGF	 & 67.98     & 73.11    & 75.54	    & 76.69	    & 80.3  \\
    \hline   
                            LPT	         & 63.96 	& 68.73 	& 71.00 	& 72.30 	& 74.20  \\
    \midrule
    \multirow{2}{*}{Method} & \multicolumn{5}{c}{ID Acc. ($\%$) (ID dataset: FOOD-101)} \\
    \cline{2-6}
                                         &2-shot    &4-shot     &8-shot     &16-shot    & All-shot  \\
    \hline
                            FFT	         & 27.06     & 43.5 	& 53.07 	& 67.96 	& 89.24   \\
    \rowcolor{gray!25}      FFT + DSGF	 & 41.05     & 53.44 	& 65.66 	& 72.94 	& 89.52    \\
    \hline   
                            VAT	         & 48.99     & 59.92 	& 69.57 	& 75.24 	& 87.57    \\
    \rowcolor{gray!25}      VAT + DSGF	 & 50.95     & 61.82 	& 69.92 	& 75.40 	& 86.98   \\
    \hline   
                            VPT	         & 50.48     & 66.20 	& 72.84 	& 77.34 	& 87.87    \\
    \rowcolor{gray!25}      VPT + DSGF	 & 52.02     & 66.83 	& 73.68 	& 77.54 	& 88.41  \\
    \hline   
                            LPT	         & 47.89 	& 58.35 	& 64.97 	& 68.19 	& 80.38 \\
    \midrule
    \multirow{2}{*}{Method} & \multicolumn{5}{c}{ID Acc. ($\%$) (ID dataset: CIFAR-100)} \\
    \cline{2-6}
                                         &2-shot    &4-shot     &8-shot     &16-shot    & All-shot  \\
    \hline   
                            FFT	         & 21.30 	 & 29.85    & 48.64     & 59.07 	& 84.50   \\
    \rowcolor{gray!25}    FFT + DSGF	 & 26.35 	 & 40.33 	& 56.05 	&61.25	    & 84.82    \\
    \hline   
                            VAT	         & 29.35 	 & 40.90 	& 55.76 	& 66.37 	& 80.61    \\
    \rowcolor{gray!25}    VAT + DSGF	 & 30.13 	 & 41.62 	& 55.41 	& 66.30 	& 80.94   \\
    \hline   
                            VPT	         & 31.20 	 & 54.78    & 69.11     & 72.92 	& 82.65    \\
    \rowcolor{gray!25}    VPT + DSGF	 & 31.56 	 & 54.80 	& 69.17     & 72.68 	& 82.60   \\
    \hline   
                            LPT	         & 29.70 	 &  37.40 	& 45.89     & 49.80     & 61.10   \\
    \hline
  \end{tabular}
\vspace{-0.6cm}
\end{table}

The main result is three-fold: \textbf{(i).} DSGF improves the OOD detection performance across various shot settings, particularly in the 2-shot and 4-shot scenarios. For example, in the 2-shot and 4-shot experiments on Imagenet-1k, DSGF increases the average AUROC scores of six OOD detection methods by $7.58\%~(73.56\%~vs.~81.14\%)$ and $5.35\%~(80.65\%~vs.~86.00\%)$, respectively. The OOD detection task also requires models to detect OOD samples without affecting ID accuracy. Therefore, we also display the ID accuracy for three ID datasets in Table~\ref{Table4}. The results indicate that DSGF can also improve the ID accuracy in most experiments, particularly in few-shot settings. For example, in the 2-shot experiment on the FOOD-101 dataset, DSGF increases the ID Acc score from $27.06\%$ to $41.05\%$. \textbf{(ii).} DSGF improves performance across different fine-tuning paradigms. Surprisingly, DSGF achieves competitive performance with just a few images for fine-tuning, particularly on the FOOD-101 dataset. Furthermore, our method exhibits a more significant improvement in the fully fine-tuning paradigm. For instance, in the 2-shot setting on the FOOD-101 dataset, the AUROC score increases on average by $6.28\%$, reaching $98.56\%$. Our method dramatically narrows the OOD detection performance gap between fully fine-tuning and efficient fine-tuning. It is worth noting that DSGF can further enhance the OOD detection performance of VAT and VPT, even though they are already performing quite well. \textbf{(iii).} DSGF enhances the performance of various post-hoc OOD detection methods. We provide a detailed breakdown of the performance improvements achieved by DSGF for each OOD method in Appendix~\ref{AppendixC.3}.

Comparing with the state-of-the-art out-of-distribution detection methods~\cite{miyai2023locoop} based on visual-language models is meaningful, but it is not the primary focus of this paper. We will discuss the comparison results in Appendix~\ref{AppendixC.2}.

\begin{figure*}[t]
\vspace{-0.2cm}
  \centering
    \includegraphics[width=0.95\linewidth]{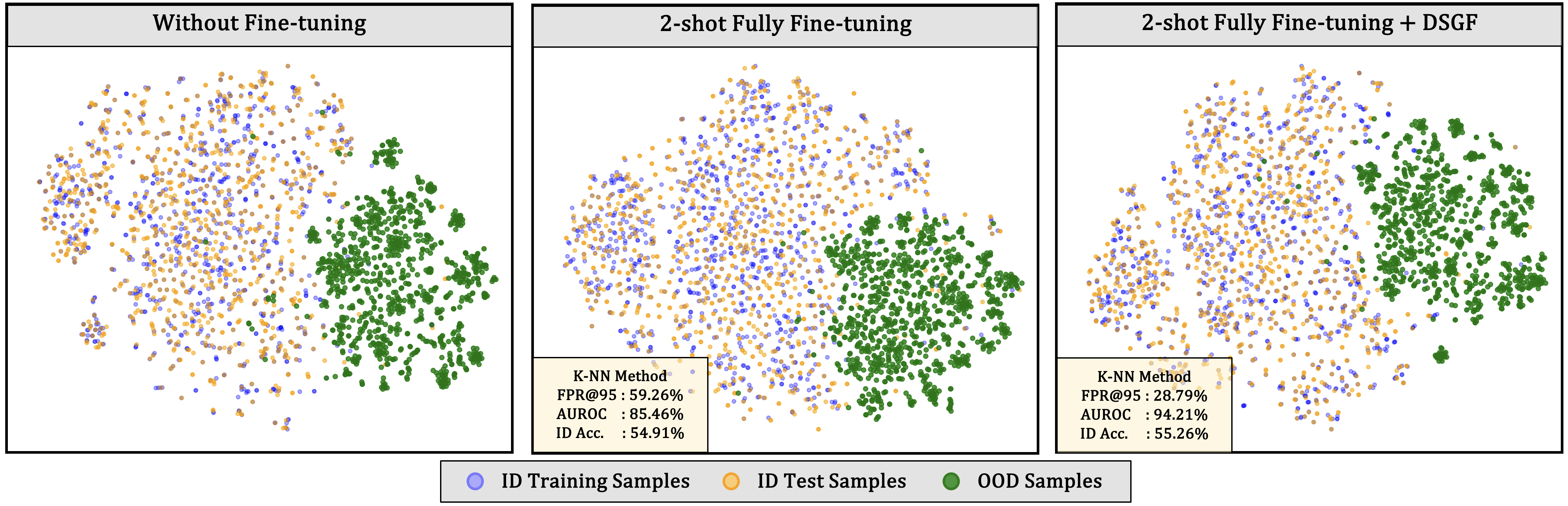}\vspace{-0.3cm}
   \caption{The t-sne visualization on Imagenet-1k (ID) and iNaturalist (OOD). We use the output features from the last layer of the Transformer to visualize the distribution of ID and OOD samples.vspace{-0.5cm}}
   \label{Fig5}
\end{figure*}

\section{Discussions and Analysis}
\label{Discussions and Analysis}
\noindent  \textbf{Feature Distribution Visualization.} 
We visualized the feature distribution of ID and OOD samples using t-SNE~\cite{van2008visualizing}. Fig.~\ref{Fig5} shows that the original pre-trained model without fine-tuning can well separate ID and OOD samples. However, this separability becomes blurred after the few-shot fine-tuning.
Our DSGF makes the boundary of ID and OOD samples clear again with the help of general knowledge from the pre-trained model, which is supported by both the visualization figures and the AUROC scores (85.46\% and 94.21\%). Our DSGF also keeps the ID accuracy of the fine-tuned model.

\begin{figure}[t]
\vspace{-0.2cm}
\centering
    \includegraphics[width=1\linewidth]{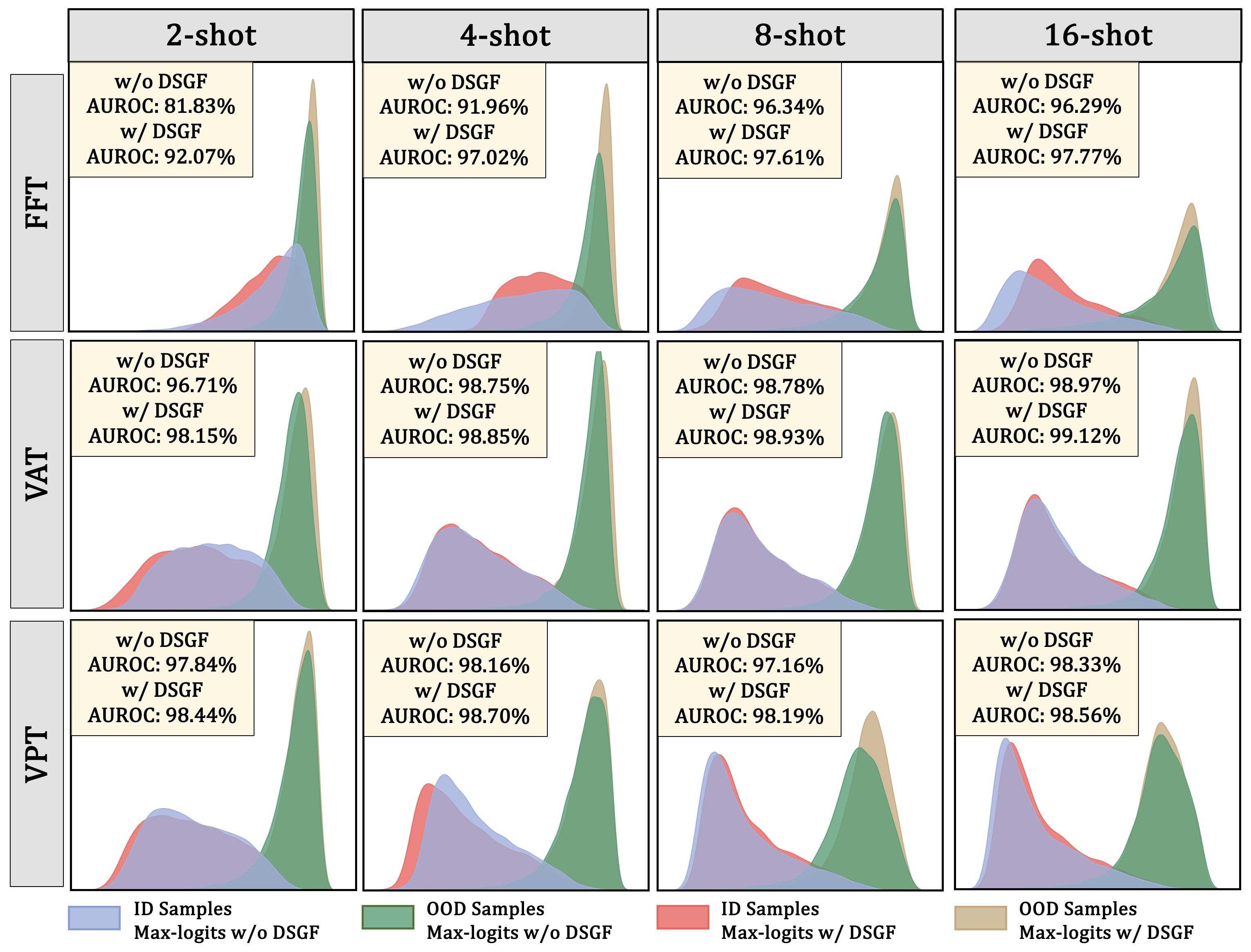}\vspace{-0.2cm}
   \caption{Uncertainty scores distribution of Imagenet-1k (ID) and iNaturalist (OOD). Each row and column of sub-figures represents different fine-tuning paradigms and shot settings. We estimate these scores using the max-logits-based method.\vspace{-0.5cm}}
   \label{Fig6}
\end{figure}
\noindent  \textbf{Uncertainty Distribution Visualization.} We visualize the distribution of uncertainty scores based on maximum logit values in Fig.~\ref{Fig6}. We find that uncertainty scores of ID samples decrease as the number of shots increases for all fine-tuning methods, which is reasonable since more training samples make the model more confident. However, uncertainty scores of OOD samples also decrease in FFT, which is not optimal for separating ID and OOD samples. PEFT methods including VPT and VAT have relatively stable uncertainty scores for OOD samples under different shots, which makes them have better OOD detection performance than FFT. The reason is that these PEFT methods freeze the pre-trained backbone, which provides useful general knowledge to distinguish OOD samples. Fig.~\ref{Fig6} shows that the brown distribution is higher than the green distribution, which means our DSGF could improve the uncertainty score of OOD samples. This is achieved by introducing the general knowledge from the pre-trained model. The higher AUROC scores also show the effectiveness of our DSGF.

\begin{figure}[t]
\vspace{-0.2cm}
  \centering
    \includegraphics[width=1\linewidth]{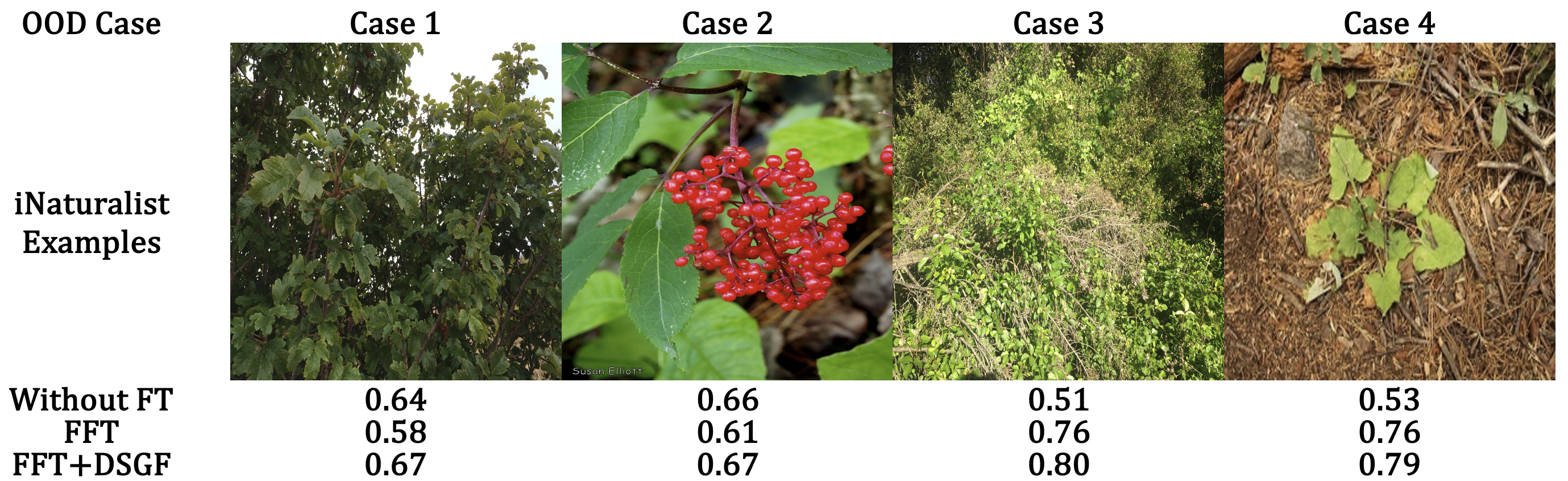}\vspace{-0.2cm}
   \caption{Uncertainty scores of OOD samples from iNaturalist dataset. We extract features using the original pre-trained model, the fully fine-tuned model, and the fusion model, and then assess the uncertainty scores of OOD samples using k-NN.\vspace{-0.5cm}}
   \label{Fig7}
\end{figure}

\noindent \textbf{Case Analysis on iNaturalist Dataset.}
We use the k-NN method to estimate uncertainty scores for several OOD samples from the iNaturalist dataset, as shown in Fig.~\ref{Fig7}. The fine-tuned model assigns lower uncertainty scores for Case 1/2 because of losing the general knowledge of the pre-trained model, and the original pre-trained model assigns low uncertainty scores for Case 3/4 without the domain-specific knowledge. In all cases, our DSGF method, which fuses the domain-specific and general knowledge, assigns the highest uncertainty scores for these OOD samples.

\section{Conclusion}
\label{Conclusion}

This paper pioneers formulating a novel and critical task: few-shot out-of-distribution detection. We construct a comprehensive FS-OOD detection benchmark and include six OOD detection baselines. Our benchmark rigorously examines a variety of fine-tuning paradigms, including traditional approaches like fully fine-tuning and linear probing tuning, as well as parameter-efficient fine-tuning methods, such as visual adapter tuning and visual prompt tuning. Furthermore, we introduce a universal approach, DSGF, which effectively merges the original pre-trained model's general knowledge with the fine-tuned model's domain-specific knowledge. Experimental results demonstrate that our method enhances FS-OOD detection across different fine-tuning strategies, shot settings, and OOD detection methods. We hope our work paves the path for research communities on the FS-OOD detection challenge.






\bibliography{example_paper}
\bibliographystyle{icml2024}

\newpage
\appendix
\onecolumn
\setcounter{page}{1} %

\section{Additional Experiments Material}
\label{AppendixA}

\subsection{Datasets}
\label{AppendixA.1}
In the context of our study, we utilize a diverse array of in-distribution (ID) and out-of-distribution (OOD) datasets to rigorously evaluate the performance of various detection methods. 

\textbf{ID Datasets.} \textit{Imagenet-1k}~\cite{deng2009imagenet} is a large-scale image classification dataset comprising 1,000 categories with over one million training images. 

\noindent \textit{FOOD-101}\cite{parkhi2012cats} encompasses 101 common food categories, containing 101,000 images.

\noindent \textit{Oxford-PET}~\cite{parkhi2012cats} contains 37 pet categories, with approximately 200 images per category. 

\noindent \textit{CIFAR-100}~\cite{krizhevsky2009learning} consists of 60,000 32x32 color images and 100 classes. There are 500 training images and 100 testing images per class.

\noindent  
\textbf{OOD Datasets.} \textit{iNaturalist}~\cite{van2018inaturalist} features 13 supercategories and 5,089 fine-grained categories, covering various organisms, including plants, insects, birds, mammals, and more. We use a subset of 110 plant categories that do not overlap with Imagenet-1k.

\noindent  \textit{SUN}~\cite{xiao2010sun} comprises 899 categories, encompassing indoor, urban, and natural scenes, both with and without human presence. We utilize a subset of 50 natural object categories not found in Imagenet-1k.

\noindent  \textit{Places}~\cite{zhou2017places} includes photos labeled with semantic scene categories from three major classes: indoor, natural, and urban scenes. Our subset consists of 50 categories sampled from those not present in Imagenet-1k.

\noindent  \textit{Texture}~\cite{cimpoi2014describing} includes images of textures and abstract patterns. As there are no overlapping categories with Imagenet-1k, we use the entire dataset as presented in~\cite{huang2021mos}.

\noindent \textit{Tiny-imagenet}~\cite{le2015tiny} contains 100,000 images of 200 classes (500 for each class) downsized to 64×64 colored images. Each class is represented by 500 training, 50 validation, and 50 test images. Notably, we resize these images from 64x64 to 32x32 to align with the resolution of CIFAR-100. This step is crucial to avoid the model distinguishing OOD samples based solely on image resolution, which would be an unsound basis for OOD detection.

This diverse selection of datasets, carefully curated to ensure relevance and non-overlap where necessary, provides a robust platform for evaluating the effectiveness of OOD detection methodologies in a wide range of scenarios.

\subsection{OOD Scores Estimation}
\label{AppendixA.2}
In Section 2.1 of our paper, we provide an initial overview of the out-of-distribution (OOD) detection methods that have been replicated and analyzed in our research. These methods predominantly rely on post-processing techniques that transform model logits into uncertainty scores. This transformation process is mathematically represented through Eq.~\ref{eq6} to~\ref{eq10}. 
\begin{equation}
  Energy=-log\sum\nolimits_{j=1}^K z_j/T
  \label{eq6}
\end{equation}
\begin{equation}
  Entropy=Entropy(\frac{e^{z_i}/T} {\sum\nolimits_{j=1}^K e^{z_j}/T})
  \label{eq7}
\end{equation}
\begin{equation}
  Variance=-Variance(\frac{e^{z_i}/T} {\sum\nolimits_{j=1}^K e^{z_j}/T})
  \label{eq8}
\end{equation}
\begin{equation}
  MSP=-Max(\frac{e^{z_i}/T} {\sum\nolimits_{j=1}^K e^{z_j}/T})
  \label{eq9}
\end{equation}
\begin{equation}
  Max-Logits=-Max(\frac{{z_i}/T} {\sum\nolimits_{j=1}^K {z_j}/T})
  \label{eq10}
\end{equation}
where $z_i$ denotes the logits of class $i$, $K$ denotes the number of classes in ID dataset, and $T$ denotes the temperature scaling factor. In this paper, we used $T=1$ as the default.

Further delving into the computation of OOD scores, we utilize Algorithm~\ref{Algorithm 1} to illustrate the procedure for employing the k-nearest neighbors (k-NN) method. A noteworthy aspect of the k-NN approach is its ability to perform training-free computations. It achieves this by calculating the similarity between a given test image and the images in the training dataset, thereby bypassing the need for additional model training. 

\begin{algorithm}[t]
\caption{OOD Detection with k-NN Feature Matching}\label{alg:cap}
\label{Algorithm 1}
\begin{algorithmic}
    \STATE {\bfseries Input:} training images $X^{train}$, test images $X^{test}$, pre-trained model $M_o$, fine-tuned model $M_{ft}$.
    \STATE {\bfseries do:}
    \begin{itemize}
        \item Collect feature vectors $M_o(X^{train})$ or $M_{ft}(X^{train})$ for all training images $X^{train}$ from $M_o$ or $M_{ft}$ 
        \item Collect feature vectors $M_o(X_i^{test})$ or $M_{ft}(X_i^{test})$ for each test image $X_i^{test}$ from $M_o$ or $M_{ft}$ 
        \item Calculate the cosine similarity between each test image's and training images' features. Then, take the negative of the maximum value as the OOD score.
    \end{itemize}
    \STATE {\bfseries Output:} OOD score of a test image can be formulated as follows:
\begin{equation}
  k-NN=-Max(Similarity(M_*(X_i^{test}),M_*(X^{train}))
  \label{eq11}
\end{equation}
\end{algorithmic}
\end{algorithm}


\subsection{Evaluation Metrics}
\label{AppendixA.3}
\label{sec:rationale}

\textit{FPR@TPR95} can be interpreted as the probability that a negative (out-of-distribution) example is misclassified as positive (in-distribution) when the true positive rate (TPR) is as high as $95\%$. The true positive rate can be computed by $TPR=TP/(TP+FN)$, where $TP$ and $FN$ denote true positives and false negatives, respectively. The false positive rate (FPR) can be computed by $FPR=FP/(FP+TN)$, where FP and TN denote false positives and true negatives, respectively.

\textit{AUROC.} By treating in-distribution data as positive and out-of-distribution data as negative, various thresholds can be applied to generate a range of true positive rates (TPR) and false-positive rates (FPR). From these rates, we can calculate AUROC.

\section{DSGF method}
\label{AppendixB}

\textbf{Algorithm.}
Upon completing the fine-tuning process, the optimal checkpoint is selected to represent the fine-tuned model, which forms the basis for developing our DSGF method. The procedural steps of DSGF are succinctly outlined in Algorithm \ref{Algorithm 2}. In the context of employing the k-nearest neighbors (k-NN) method, we explore three distinct operational approaches:

\begin{itemize}
    \item Concatenation Approach: This involves concatenating the features extracted from the pre-trained model with those from the fine-tuned model. Subsequent to this concatenation, the combined feature set is normalized. The k-NN method is then applied to these normalized, concatenated features to compute the k-NN scores.
    
    \item Separate Normalization and Concatenation Approach: Here, we first normalize the features from the pre-trained model and the fine-tuned model independently. These normalized feature sets are then concatenated, followed by the application of the k-NN method to derive the k-NN scores.
    
    \item Separate Normalization and Score Summation Approach: In this method, the features from the pre-trained and fine-tuned models are normalized independently. The k-NN method is then applied to each set of normalized features to obtain respective k-NN scores, which are subsequently summed to yield the final score.
\end{itemize}

Interestingly, our findings indicate that the end results of these three operational methods are identical in their performance metrics.

\begin{algorithm}
    \caption{Domain-Specific and General Knowledge Fusion for Few-Shot OOD Detection}
    \label{Algorithm 2}
\begin{algorithmic}
    \STATE {\bfseries Input:} training images $X^{train}$, training labels $Y^{train}$, test images $X^{test}$, pre-trained model $M_o$, fine-tuned model $M_{ft}$.
    \STATE {\bfseries For each $X_i^{train}$ in $X^{train}$:}
    \begin{itemize}
        \item Collect feature vectors $M_o(X_i^{train})$ and $M_{ft}(X_i^{train})$ from $M_o$ and $M_{ft}$ 
        \item Concatenate two feature vectors to $[M_o(X_i^{train}), M_{ft}(X_i^{train})] \in \mathbb{R}^{2d}$
        \item Train a linear classifier using the concatenated feature $[M_o(X_i^{train}), M_{ft}(X_i^{train})]$ and label $Y_i^{train}$
    \end{itemize}
    \STATE {\bfseries Output:} The logits vector $z_i$ of each image in $X^{test}$, which can be converted to the OOD score by Eq.\ref{eq6} to Eq.\ref{eq10}
\end{algorithmic}
\end{algorithm}

\begin{figure*}[t]
\vspace{-0.2cm}
  \centering
    \includegraphics[width=0.95\linewidth]{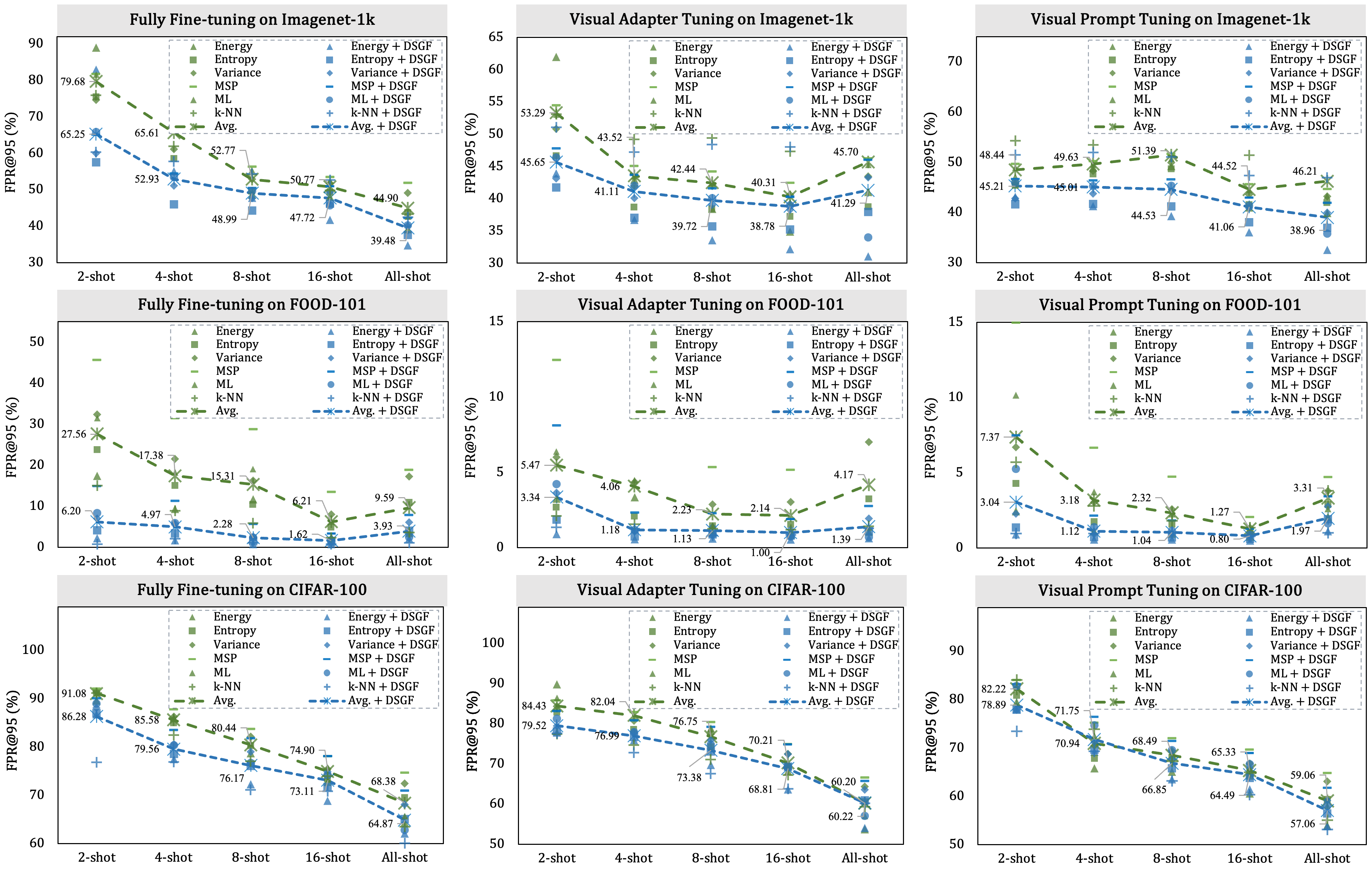}\vspace{-0.2cm}
  \caption{Main Results of Few-Shot OOD detection on different tuning paradigms. Overall, our method achieves more performance gains in few-shot settings. `Avg.' represents the arithmetic average of six out-of-distribution score evaluation methods, and `DSGF' denotes deploying our method.\vspace{-0.5cm}}
  \label{Fig8}
\end{figure*}

\noindent \textbf{Additional Implementation Details.} To comprehensively understand our methodology, we delineate the hyper-parameter settings for STAGE 1 and STAGE 2 in Table~\ref{Table5}. Notably, all experiments conducted within the scope of this study are feasible on a single GPU setup, ensuring accessibility and reproducibility of our experimental framework within the constraints of typical computational resources.

\begin{table*}[t]
\vspace{-0.3cm}
\centering
  \caption{Hyper-parameters setting for four ID datasets on different settings}
  \label{Table5}
    \fontsize{7}{8}\selectfont
  \begin{tabular}{cccc|ccc}
                                    & \multicolumn{3}{c}{STAGE 1}                            &\multicolumn{3}{c}{STAGE 2}\\
    \midrule
                                    & \multicolumn{6}{c}{Imagenet-1k}\\
    \cline{2-7}
    {} & Learning Rate & Weight Decay & Batch Size & Learning Rate & Weight Decay & Batch Size	 \\
    \cline{2-7}
                FFT  & 0.0001                                    & 0.01   & 64        & 0.01 / 0.01 / 0.1 / 0.1 / 0.001 & 0.0001 / 0.0001/ 0.0001 / 0 / 0        & 32  \\
                VAT  & 0.01 / 0.001 / 0.001 / 0.001 / 0.001      & 0.001  & 128       & 0.1                             & 0                                 & 32 \\
                VPT  & 2.5 / 1.25 / 1.25 / 2.5 / 1.25            & 0.001  & 128       & 0.1 / 0.1 / 0.1 / 0.1 / 0.01    & 0.001 / 0.001 / 0.001 / 0.001 / 0 & 32 \\
                LPT  & 2.5                                       & 0.0001 & 512       & - & - & -  \\
    \midrule
                                    & \multicolumn{6}{c}{FOOD-101} \\ 
    \cline{2-7}
    {} & Learning Rate & Weight Decay & Batch Size & Learning Rate & Weight Decay & Batch Size	 \\
    \cline{2-7}
                FFT  & 0.001 / 0.001 / 0.0001 / 0.0001 / 0.0001  & 0.01   & 64       & 0.1                             & 0                                 & 32  \\
                VAT  & 0.01                                      & 0.01   & 128      & 0.1                             & 0                                 & 32 \\
                VPT  & 3.75 / 3.75 / 2.5 / 2.5 / 1.25            & 0.001  & 128      & 0.1                             & 0.01 / 0.01 / 0.01 / 0.01 / 0.001 & 32 \\
                LPT  & 2.5                                       & 0.0001 & 512      & - & - & -  \\
    \midrule   
                                    & \multicolumn{6}{c}{Oxford-PETS}  \\ 
    \cline{2-7}
    {} & Learning Rate & Weight Decay & Batch Size & Learning Rate & Weight Decay & Batch Size	 \\
    \cline{2-7}
                FFT  & 0.0001                                    & 0.01   & 64         & 0.1                             & 0                                  & 32  \\
                VAT  & 0.01                                      & 0.01   & 128        & 0.1                             & 0                                  & 32 \\
                VPT  & 3.75 / 2.5 / 2.5 / 2.5 / 1.25             & 0.001  & 128        & 0.1                             & 0.1 / 0.1 / 0.1 / 0.1 / 0.01       & 32 \\
                LPT  & 2.5                                       & 0.0001 & 512        & - & - & -  \\
    \midrule   
                                    & \multicolumn{6}{c}{CIFAR-100}  \\ 
    \cline{2-7}
    {} & Learning Rate & Weight Decay & Batch Size & Learning Rate & Weight Decay & Batch Size	 \\
    \cline{2-7}
                FFT  & 0.001 / 0.001 / 0.0001 / 0.0001 / 0.0001  & 0.01   & 64        & 0.1                             & 0                                  & 32  \\
                VAT  & 0.01 / 0.01 / 0.01 / 0.01 / 0.001         & 0.01   & 128       & 0.1                             & 0                                  & 32 \\
                VPT  & 3.75 / 2.5 / 2.5 / 1.25 / 3.75            & 0.001  & 128       & 0.1                             & 0.1 / 0.01 / 0.01 / 0.01 / 0.01    & 32 \\
                LPT  & 2.5                                       & 0.0001 & 512       & - & - & -  \\
    \midrule   
    \multicolumn{7}{c}{*/*/*/*/* The five values correspond to the settings of 2/4/8/16/All shot respectively. A single value applies to 2/4/8/16/All shot settings.}\\
\end{tabular}
\vspace{-0.3cm}
\end{table*}

\section{Results}
\label{AppendixC}
\subsection{Visualization of the FPR@95 Scores}
\label{AppendixC.1}

In our study, we conducted a comprehensive comparison of the False Positive Rate at $95\%$ True Positive Rate (FPR@95) scores between our DSGF method and various baseline methods, as shown in Fig.~\ref{Fig8}. Specifically, the green markers in these subfigures denote the FPR@95 scores obtained using the six original out-of-distribution (OOD) detection methods. In contrast, the blue markers signify the results achieved subsequent to the implementation of our DSGF method.

\subsection{Comparison with State-of-the-Art CLIP-Based Approaches}
\label{AppendixC.2}

Recently, some methods based on vision-language models~\cite{miyai2023locoop} aim to address the problem of FS-OOD detection by leveraging text-image matching and have achieved state-of-the-art performance. LoCoOp~\cite{miyai2023locoop} utilizes out-of-distribution (OOD) regularization techniques to penalize regions in images that are irrelevant to in-distribution (ID) categories, surpassing its baseline methods~\cite{zhou2022conditional}. We have compared these methods' FPR@95 and AUROC scores in Table~\ref{Table_C.2.1}. Additionally, the in-distribution accuracy of these approaches is compared in Table~\ref{Table_C.2.2}.

The results indicate that our method is slightly inferior to the vision-language model-based approaches regarding FPR@95 and AUROC scores. We believe this may be due to the disparity in the scale of the pretraining models. CLIP was trained on nearly 400 million image-text pairs, while our pretraining model only utilized 14 million images, which is an order of magnitude difference. Despite this, we achieved FPR@95 and AUROC scores comparable to those of $CoCo_{MCM}$ and $CoCo_{GL}$~\cite{zhou2022conditional}, and significantly higher in-distribution accuracy than LoCoOp~\cite{miyai2023locoop} ($71.7\% vs. 76.69\%$). Furthermore, while the addition of OOD regularization loss in ~\cite{miyai2023locoop} improved performance in out-of-distribution detection, it also compromised the model's accuracy in-distribution. We argue that a robust out-of-distribution detection method should not do so. In contrast, our method enhances the FPR@95, AUROC scores, and in-distribution (ID) accuracy in most settings.

\begin{table*}[h]
\vspace{-0.3cm}
  \centering
  \caption{Comparison with state-of-the-art vision-language model-based methods in terms of FPR@95 and AUROC scores. The training set is ImageNet-1k under the 16-shot setting.}
  \label{Table_C.2.1}
    \fontsize{8}{10}\selectfont
  \begin{tabular}{cccccc}
    \hline
    \multirow{2}{*}{Method}
    & \multicolumn{5}{c}{\textbf{FPR@95 $\downarrow$ / AUROC $\uparrow$ $(\%)$}} \\
    \cline{2-6}
    &{iNaturalist} & {SUN} & {Places} & {Texture} & {Average}	 \\
    \hline
            $CoOp_{MCM}$~\cite{zhou2022conditional}            & 28.00 / 94.43        & 36.95 / 92.29          & 43.03 / 89.74         & \textbf{39.33} / 91.24 & 36.83 / 91.93 \\
            $CoOp_{GL}$~\cite{zhou2022conditional}             & 14.60 / 96.62        & 28.48 / 92.65          & 36.49 / 89.98         & 43.13 / 88.03         & 30.68 / 91.82 \\
            $LoCoOp_{MCM}$~\cite{miyai2023locoop}          & 23.06 / 95.45        & 32.70 / 93.35          & 39.92 / 90.64         & 40.23 / \textbf{91.32} & 33.98 / 92.69 \\
            $LoCoOp_{GL}$~\cite{miyai2023locoop}           & 16.05 / 96.86        & \textbf{23.44 / 95.07} & \textbf{32.87 / 91.98} & 42.28 / 90.19          & \textbf{28.66 / 93.53} \\
    \midrule 
            $DSGF_{Energy}(VPT)$    & \textbf{3.39 / 99.18} & 42.31 / 89.80          & 50.34 / 86.49          & 48.23 / 89.90        & 36.07 / 91.34 \\ 
    \hline
  \end{tabular}
  \vspace{-0.3cm}
\end{table*}

\begin{table*}[h]
\vspace{-0.3cm}
  \centering
  \caption{Comparison with state-of-the-art vision-language model-based methods in terms of in-distribution accuracy. The training set is ImageNet-1k under the 16-shot setting.}
  \label{Table_C.2.2}
    \fontsize{8}{10}\selectfont
  \begin{tabular}{cc}
    \hline
    {Method} &\textbf{In-distribution Accuracy $\uparrow$ $(\%)$}	 \\
    \hline
            $CoOp$~\cite{zhou2022conditional}                  & 72.10     \\
            $LoCoOp$~\cite{miyai2023locoop}                & 71.70     \\
    \hline
            $DSGF_{Energy}(VPT)$    & \textbf{76.69}     \\ 
    \hline
  \end{tabular}
\vspace{-0.3cm}
\end{table*}

\subsection{Benchmark Details}
\label{AppendixC.3}
In our endeavor to establish a comprehensive benchmark, we meticulously report extensive results encompassing each pairing of in-distribution (ID) and out-of-distribution (OOD) datasets. This detailed reporting includes an array of variables, such as different tuning paradigms, shot settings, OOD detection methods, and a comparative analysis with our DSGF method. 

To facilitate a thorough and transparent evaluation, we present the FPR@95 and AUROC scores across a series of tables, specifically from Table~\ref{Table6} to Table~\ref{Table18}. These tables collectively offer a granular view of the performance metrics under various conditions and configurations, thereby providing a nuanced understanding of the efficacy of each method.

Additionally, we dedicate Table~\ref{Table19} to the presentation of in-distribution accuracy metrics. This inclusion ensures a holistic assessment of the models' performance, not just in terms of their ability to detect OOD instances, but also in accurately identifying ID instances. 


\begin{table*}[h]
  \centering
  \caption{Comparison of Baseline and Our DSGF. ID and OOD datasets are Imagenet-1k and SUN.}
  \label{Table6}
    \fontsize{7}{8}\selectfont
\begin{tabular}{ccccccc}
    \midrule
    \multirow{2}{*}{2-shot}
    & \multicolumn{6}{c}{FPR@95 $\downarrow$/AUROC $\uparrow$ $(\%)$} \\
    \cline{2-7}
    {} & Energy & Entropy & Variance & MSP	& Max-Logits & k-NN	 \\
    \hline
                       FFT         & 91.70/64.25  & 83.10/70.07 & 82.66/69.64 & 87.07/67.13 & 87.10/67.20 & 83.54/71.11 \\
   \rowcolor{gray!25}   FFT + DSGF  & 84.25/73.32  & 67.31/78.52 & 70.63/77.35 & 75.64/76.30 & 75.61/76.29 & 74.29/77.34 \\
   \cline{1-7}     
                       VAT         & 61.36/83.16  & 54.53/83.52 & 60.37/82.12 & 64.22/81.33 & 61.41/82.65 & 66.56/80.76 \\
   \rowcolor{gray!25}  VAT + DSGF  & 47.11/86.87  & 48.03/85.94 & 51.92/84.89 & 58.74/83.76 & 56.65/84.47 & 61.27/82.78 \\
   \cline{1-7}     
                       VPT         & 44.56/85.46  & 47.01/84.52 & 52.20/83.58 & 58.96/82.58 & 57.37/82.99 & 67.39/81.34 \\
   \rowcolor{gray!25}  VPT + DSGF  & 46.63/86.23  & 47.86/85.60 & 52.16/84.83 & 58.07/84.00 & 56.79/84.41 & 62.24/83.35 \\
\end{tabular}
\end{table*}


\begin{table*}[t]
  \centering
  \caption{Comparison of Baseline and Our DSGF. ID and OOD datasets are Imagenet-1k and iNaturalist.}
  \label{Table7}
    \fontsize{7}{8}\selectfont
  \begin{tabular}{ccccccc}
    \midrule
    \multirow{2}{*}{2-shot}
    & \multicolumn{6}{c}{FPR@95 $\downarrow$/AUROC $\uparrow$ $(\%)$} \\
    \cline{2-7}
    {} & Energy & Entropy & Variance & MSP	& Max-Logits & k-NN	 \\
    \hline   
                       FFT         & 85.62/75.41 & 57.74/86.66 &  56.89/86.11  & 70.80/81.69 & 70.50/81.83  & 59.26/85.46   \\
   \rowcolor{gray!25}  FFT + DSGF  & 66.19/88.12 & 23.58/94.51 &  28.55/93.33  & 37.85/91.75 & 36.96/92.07  & 28.79/94.21   \\
   \cline{2-7}     
                       VAT         & 35.95/94.77 & 10.28/97.79 &  15.20/97.01  & 20.65/96.21 & 18.63/96.71  & 16.84/96.62   \\
   \rowcolor{gray!25}  VAT + DSGF  & 5.42/98.46 & 4.73/98.69 &  6.23/98.56  & 10.71/97.95 & 9.44/98.15  & 17.69/96.39   \\
   \cline{2-7}     
                       VPT         & 6.79/98.38 & 6.17/98.54 &  7.52/98.37  & 11.86/97.71 & 11.03/97.84  & 12.39/97.71   \\
   \rowcolor{gray!25}  VPT + DSGF  & 4.27/98.93 & 3.94/99.03 &  5.06/98.87  & 8.36/98.32 & 7.76/98.44  & 15.82/97.01   \\
\end{tabular}
\end{table*}


\begin{table*}[t]
  \centering
  \caption{Comparison of Baseline and Our DSGF. ID and OOD datasets are Imagenet-1k and Places.}
  \label{Table8}
    \fontsize{7}{8}\selectfont
  \begin{tabular}{ccccccc}
    \midrule
    \multirow{2}{*}{2-shot}
    & \multicolumn{6}{c}{FPR@95 $\downarrow$/AUROC $\uparrow$ $(\%)$} \\
    \cline{2-7}
    {} & Energy & Entropy & Variance & MSP	& Max-Logits & k-NN	 \\
    \hline    
                       FFT         & 92.35/62.56 & 88.09/65.43 & 87.68/65.42  & 89.42/64.20 & 89.37/64.28  & 86.12/68.48   \\
   \rowcolor{gray!25}  FFT + DSGF  & 88.74/69.38 & 77.37/74.02 & 78.66/73.50  & 81.02/72.97 & 81.33/72.73  & 78.00/75.10   \\
   \cline{2-7}     
                       VAT         & 71.97/79.47 & 66.18/79.93 & 69.81/79.03  & 72.10/78.48 & 70.23/79.35  & 71.18/78.86   \\
   \rowcolor{gray!25}  VAT + DSGF  & 62.23/82.25 & 61.36/81.70 & 63.23/81.14  & 67.23/80.49 & 65.87/81.01  & 68.61/79.87   \\
   \cline{2-7}     
                       VPT         & 57.23/81.70 & 58.60/81.08 & 61.98/80.54  & 67.02/79.90 & 66.13/80.17  & 71.04/79.27   \\
   \rowcolor{gray!25}  VPT + DSGF  & 60.08/82.07 & 59.80/81.80 & 62.31/81.47  & 66.25/80.99 & 65.47/81.25  & 68.25/80.59   \\
\end{tabular}
\end{table*}

 
\begin{table*}[t]
  \centering
  \caption{Comparison of Baseline and Our DSGF. ID and OOD datasets are Imagenet-1k and Texture.}
  \label{Table9}
    \fontsize{7}{8}\selectfont
  \begin{tabular}{ccccccc}
    \midrule
    \multirow{2}{*}{2-shot}
    & \multicolumn{6}{c}{FPR@95 $\downarrow$/AUROC $\uparrow$ $(\%)$} \\
    \cline{2-7}
    {} & Energy & Entropy & Variance & MSP	& Max-Logits & k-NN	 \\
    \hline   
                       FFT         & 86.19/74.20 & 72.62/81.15 & 71.17/80.43  & 79.56/76.72 & 79.29/76.93  & 74.45/79.12  \\
   \rowcolor{gray!25}  FFT + DSGF  & 92.41/74.34 & 61.72/85.17 & 62.07/84.56  & 65.78/83.57 & 69.04/83.15  & 60.14/85.70  \\
   \cline{2-7}     
                       VAT         & 78.46/84.15 & 55.43/87.07 & 57.80/86.06  & 60.57/85.28 & 61.67/85.97  & 57.50/86.67  \\
   \rowcolor{gray!25}  VAT + DSGF  & 60.96/87.89 & 52.78/88.90 & 51.03/88.74  & 54.41/88.02 & 53.46/88.42  & 56.44/87.12  \\
   \cline{2-7}       
                       VPT         & 79.08/82.68 & 69.59/85.84 & 61.01/86.87  & 60.50/86.70 & 60.89/86.72  & 66.17/86.04  \\
   \rowcolor{gray!25}  VPT + DSGF  & 61.45/87.93 & 54.65/88.99 & 51.40/89.05  & 53.69/88.46 & 53.35/88.70  & 59.49/87.53  \\
\end{tabular}
\end{table*}

 
\begin{table*}[t]
  \centering
  \caption{Comparison of Baseline and Our DSGF. ID and OOD datasets are FOOD-101 and SUN.}
  \label{Table10}
    \fontsize{7}{8}\selectfont
  \begin{tabular}{ccccccc}
    \midrule
    \multirow{2}{*}{2-shot}
    & \multicolumn{6}{c}{FPR@95 $\downarrow$/AUROC $\uparrow$ $(\%)$} \\
    \cline{2-7}
    {} & Energy & Entropy & Variance & MSP	& Max-Logits & k-NN	 \\
    \hline
                       FFT         & 10.42/97.29 & 17.00/95.31 & 24.67/93.06 & 37.86/90.12 & 23.37/94.22 & 10.68/97.50   \\
   \rowcolor{gray!25}  FFT + DSGF  & 0.49/99.89 & 1.66/99.66 & 3.75/99.21 & 9.97/97.97 & 4.63/99.09 & 0.28/99.90   \\
   \cline{2-7}       
                       VAT         & 1.34/99.38 & 1.16/99.66 & 3.37/99.25 & 9.10/98.29 & 3.57/99.19 & 0.68/99.84   \\
   \rowcolor{gray!25}  VAT + DSGF  & 0.21/99.89 & 0.70/99.79 & 1.38/99.59 & 4.82/98.99 & 1.84/99.53 & 0.35/99.90   \\
   \cline{2-7}       
                       VPT         & 0.64/99.81 & 1.41/99.62 & 2.80/99.29 & 9.39/98.02 & 5.51/98.79 & 2.66/99.40   \\
   \rowcolor{gray!25}  VPT + DSGF  & 0.20/99.93 & 0.37/99.89 & 0.82/99.77 & 4.78/99.00 & 2.79/99.36 & 0.31/99.92   \\
\end{tabular}
\end{table*}


\begin{table*}[t]
  \centering
  \caption{Comparison of Baseline and Our DSGF. ID and OOD datasets are FOOD-101 and iNaturalist.}
  \label{Table11}
    \fontsize{7}{8}\selectfont
  \begin{tabular}{ccccccc}
    \midrule
    \multirow{2}{*}{2-shot}
    & \multicolumn{6}{c}{FPR@95 $\downarrow$/AUROC $\uparrow$ $(\%)$} \\
    \cline{2-7}
    {} & Energy & Entropy & Variance & MSP	& Max-Logits & k-NN	 \\
    \hline
                       FFT         & 2.00/97.96 & 8.60/96.32 & 14.39/94.02  & 22.99/90.84 & 38.01/94.98  & 21.92/97.90  \\
   \rowcolor{gray!25}  FFT + DSGF  & 0.33/99.91 & 0.92/99.80 & 2.09/99.55  & 7.05/98.65 & 2.92/99.41  & 0.11/99.95  \\
   \cline{2-7}       
                       VAT         & 1.01/99.42 & 0.74/99.73 & 2.93/99.37  & 8.32/98.44 & 3.19/99.27  & 0.28/99.91  \\
   \rowcolor{gray!25}  VAT + DSGF  & 0.10/99.92 & 0.24/99.91 & 0.44/99.85  & 1.91/99.55 & 0.58/99.79  & 0.19/99.93  \\
   \cline{2-7}       
                       VPT         & 0.48/99.88 & 1.01/99.76 & 1.86/99.55  & 6.82/98.65 & 3.77/99.20  & 2.05/99.51  \\
   \rowcolor{gray!25}  VPT + DSGF  & 0.12/99.94 & 0.20/99.94 & 0.43/99.90  & 2.31/99.52 & 1.34/99.69  & 0.16/99.95  \\
\end{tabular}
\end{table*}


\begin{table*}[t]
  \centering
  \caption{Comparison of Baseline and Our DSGF. ID and OOD datasets are FOOD-101 and Places.}
  \label{Table12}
    \fontsize{7}{8}\selectfont
  \begin{tabular}{ccccccc}
    \midrule
    \multirow{2}{*}{2-shot}
    & \multicolumn{6}{c}{FPR@95 $\downarrow$/AUROC $\uparrow$ $(\%)$} \\
    \cline{2-7}
    {} & Energy & Entropy & Variance & MSP	& Max-Logits & k-NN	 \\
    \hline
                       FFT         & 14.70/96.30  & 20.97/91.80 & 28.95/91.80  & 42.26/88.79 & 27.94/92.97  & 11.06/97.35  \\
   \rowcolor{gray!25}  FFT + DSGF  & 0.83/99.78  & 2.70/98.78 & 5.40/98.78  & 13.02/97.30 & 6.70/98.68  & 0.34/99.89  \\
   \cline{2-7}       
                       VAT         & 2.25/99.20  & 1.89/98.95 & 4.89/98.95  & 11.26/97.86 & 5.20/98.93  & 0.59/99.84  \\
   \rowcolor{gray!25}  VAT + DSGF  & 0.47/99.83  & 1.04/99.37 & 2.43/99.37  & 6.45/98.60 & 2.92/99.32  & 0.41/99.88  \\
   \cline{2-7}       
                       VPT         & 1.53/99.54  & 3.13/98.51 & 5.37/98.51  & 12.94/96.90 & 8.63/97.94  & 4.19/99.03  \\
   \rowcolor{gray!25}  VPT + DSGF  & 0.58/99.82  & 1.04/99.47 & 1.85/99.47  & 6.87/98.43 & 4.79/98.91  & 0.31/99.90  \\
\end{tabular}
\end{table*}

 
\begin{table*}[t]
  \centering
  \caption{Comparison of Baseline and Our DSGF. ID and OOD datasets are FOOD-101 and Texture.}
  \label{Table13}
    \fontsize{7}{8}\selectfont
  \begin{tabular}{ccccccc}
    \midrule
    \multirow{2}{*}{2-shot}
    & \multicolumn{6}{c}{FPR@95 $\downarrow$/AUROC $\uparrow$ $(\%)$} \\
    \cline{2-7}
    {} & Energy & Entropy & Variance & MSP	& Max-Logits & k-NN	 \\
    \hline
                       FFT         & 35.89/89.42 & 42.52/86.33 & 52.61/82.78 & 64.04/79.41 & 52.29/84.55 & 29.36/91.15  \\
   \rowcolor{gray!25}  FFT + DSGF  & 7.04/98.21 & 11.21/97.10 & 16.45/95.56 & 29.49/92.86 & 19.06/95.44 & 2.25/99.36  \\
   \cline{2-7}       
                       VAT         & 8.44/97.98 & 6.88/98.37 & 12.82/97.27 & 21.08/95.71 & 13.51/97.42 & 6.84/98.45  \\
   \rowcolor{gray!25}  VAT + DSGF  & 2.75/99.21 & 5.48/98.63 & 10.34/97.63 & 19.20/95.89 & 11.56/97.69 & 4.45/98.99  \\
   \cline{2-7}       
                       VPT         & 7.07/98.25 & 11.63/97.10 & 16.74/95.73 & 30.69/92.99 & 22.73/94.81 & 13.90/96.90  \\
   \rowcolor{gray!25}  VPT + DSGF  & 2.75/99.23 & 3.83/99.01 & 6.01/98.47 & 15.96/96.58 & 12.13/97.42 & 3.09/99.19  \\
\end{tabular}
\end{table*}


\begin{table*}[t]
  \centering
  \caption{Comparison of Baseline and Our DSGF. ID and OOD datasets are Oxford-PETS and SUN.}
  \label{Table14}
    \fontsize{7}{8}\selectfont
  \begin{tabular}{ccccccc}
    \midrule
    \multirow{2}{*}{2-shot}
    & \multicolumn{6}{c}{FPR@95 $\downarrow$/AUROC $\uparrow$ $(\%)$} \\
    \cline{2-7}
    {} & Energy & Entropy & Variance & MSP	& Max-Logits & k-NN	 \\
    \hline
                       FFT         & 45.60/91.21 & 43.45/90.88 & 52.00/88.89 & 62.29/86.06 & 50.00/90.26 & 0.17/99.92  \\
   \rowcolor{gray!25}  FFT + DSGF  & 0.04/99.96 & 0.04/99.97 & 0.15/99.94 & 0.57/99.82 & 0.05/99.94 & 0.02/99.98  \\
   \cline{2-7}       
                       VAT         & 0.55/99.56 & 3.22/99.10 & 7.64/98.39 & 13.55/97.53 & 1.38/99.39 & 0.04/99.96  \\
   \rowcolor{gray!25}  VAT + DSGF  & 0.02/99.98 & 0.03/99.98 & 0.05/99.97 & 0.06/99.95 & 0.02/99.98 & 0.02/99.98  \\
   \cline{2-7}       
                       VPT         & 1.24/99.42 & 4.16/98.97 & 10.80/98.00 & 20.67/96.50 & 3.34/99.09 & 0.40/99.80  \\
   \rowcolor{gray!25}  VPT + DSGF  & 0.02/99.98 & 0.02/99.97 & 0.02/99.97 & 0.04/99.93 & 0.02/99.95 & 0.02/99.99  \\
\end{tabular}
\end{table*}


\begin{table*}[t]
  \centering
  \caption{Comparison of Baseline and Our DSGF. ID and OOD datasets are Oxford-PETS and iNaturalist.}
  \label{Table15}
    \fontsize{7}{8}\selectfont
  \begin{tabular}{ccccccc}
    \midrule
    \multirow{2}{*}{2-shot}
    & \multicolumn{6}{c}{FPR@95 $\downarrow$/AUROC $\uparrow$ $(\%)$} \\
    \cline{2-7}
    {} & Energy & Entropy & Variance & MSP	& Max-Logits & k-NN	 \\
    \hline
                       FFT         & 63.07/86.81 & 46.33/90.93 & 53.44/89.03  & 62.95/86.28 & 58.76/88.37 & 0.42/99.87   \\
   \rowcolor{gray!25}  FFT + DSGF  & 0.00/99.97 & 0.03/99.97 & 0.11/99.94  & 0.53/99.81 & 0.06/99.94 & 0.00/99.99   \\
   \cline{2-7}       
                       VAT         & 0.82/99.34 & 4.82/98.82 & 10.63/97.90  & 18.02/96.90 & 2.21/99.15 & 0.00/99.98   \\
   \rowcolor{gray!25}  VAT + DSGF  & 0.00/99.99 & 0.01/99.99 & 0.05/99.98  & 0.07/99.95 & 0.00/99.98 & 0.00/99.99   \\
   \cline{2-7}       
                       VPT         & 0.79/99.56 & 4.15/98.95 & 11.95/97.74  & 23.69/95.93 & 2.84/99.14 & 0.23/99.85   \\
   \rowcolor{gray!25}  VPT + DSGF  & 0.01/99.99 & 0.01/99.99 & 0.01/99.98  & 0.02/99.95 & 0.01/99.96 & 0.00/100.00   \\
\end{tabular}
\end{table*}


\begin{table*}[t]
  \centering
  \caption{Comparison of Baseline and Our DSGF. ID and OOD datasets are Oxford-PETS and Places.}
  \label{Table16}
    \fontsize{7}{8}\selectfont
  \begin{tabular}{ccccccc}
    \midrule
    \multirow{2}{*}{2-shot}
    & \multicolumn{6}{c}{FPR@95 $\downarrow$/AUROC $\uparrow$ $(\%)$} \\
    \cline{2-7}
    {} & Energy & Entropy & Variance & MSP	& Max-Logits & k-NN	 \\
    \hline
                       FFT         & 49.28/90.50 & 46.33/90.23 & 54.20/88.26 & 63.34/85.56 & 51.55/89.64 & 0.67/99.80  \\
   \rowcolor{gray!25}  FFT + DSGF  & 0.15/99.93 & 0.23/99.93 & 0.38/99.89 & 0.83/99.78 & 0.26/99.91 & 0.18/99.95  \\
   \cline{2-7}       
                       VAT         & 0.75/99.48 & 3.56/99.01 & 8.32/98.27 & 14.68/97.38 & 1.56/99.31 & 0.19/99.92  \\
   \rowcolor{gray!25}  VAT + DSGF  & 0.08/99.96 & 0.21/99.94 & 0.35/99.93 & 0.45/99.88 & 0.09/99.95 & 0.18/99.96  \\
   \cline{2-7}       
                       VPT         & 2.12/99.33 & 5.65/98.73 & 12.85/97.56 & 22.86/95.95 & 4.53/98.94 & 0.76/99.71  \\
   \rowcolor{gray!25}  VPT + DSGF  & 0.16/99.95 & 0.16/99.95 & 0.16/99.94 & 0.28/99.88 & 0.22/99.91 & 0.15/99.96  \\
\end{tabular}
\end{table*}

  
\begin{table*}[t]
  \centering
  \caption{Comparison of Baseline and Our DSGF. ID and OOD datasets are Oxford-PETS and Texture.}
  \label{Table17}
    \fontsize{7}{8}\selectfont
  \begin{tabular}{ccccccc}
    \midrule
    \multirow{2}{*}{2-shot}
    & \multicolumn{6}{c}{FPR@95 $\downarrow$/AUROC $\uparrow$ $(\%)$} \\
    \cline{2-7}
    {} & Energy & Entropy & Variance & MSP	& Max-Logits & k-NN	 \\
    \hline
                       FFT         & 55.59/89.80 & 48.39/90.38 & 56.44/88.52 & 65.28/85.81 & 55.50/89.46 & 0.37/99.89 \\
   \rowcolor{gray!25}  FFT + DSGF  & 0.09/99.95 & 0.12/99.95 & 0.23/99.92 & 0.55/99.81 & 0.12/99.93 & 0.07/99.97 \\
   \cline{2-7}       
                       VAT         & 0.85/99.37 & 6.01/98.66 & 12.57/97.63 & 20.37/96.55 & 2.43/99.12 & 0.07/99.95 \\
   \rowcolor{gray!25}  VAT + DSGF  & 0.04/99.97 & 0.12/99.96 & 0.34/99.94 & 0.59/99.88 & 0.04/99.95 & 0.05/99.96 \\
   \cline{2-7}       
                       VPT         & 5.96/98.49 & 18.26/96.75 & 33.95/94.16 & 48.58/91.43 & 13.17/97.63 & 0.44/99.78 \\
   \rowcolor{gray!25}  VPT + DSGF  & 0.09/99.97 & 0.09/99.96 & 0.11/99.96 & 0.20/99.89 & 0.14/99.92 & 0.12/99.95 \\
\end{tabular}
\end{table*}

\begin{table*}[t]
  \centering
  \caption{Comparison of Baseline and Our DSGF. ID and OOD datasets are CIFAR-100 and Tiny-imagenet.}
  \label{Table18}
    \fontsize{7}{8}\selectfont
  \begin{tabular}{ccccccc}
    \midrule
    \multirow{2}{*}{2-shot}
    & \multicolumn{6}{c}{FPR@95 $\downarrow$/AUROC $\uparrow$ $(\%)$} \\
    \cline{2-7}
    {} & Energy & Entropy & Variance & MSP	& Max-Logits & k-NN	 \\
    \hline
                       FFT           & 89.56/61.56 & 91.30/59.63 & 91.34/58.90 & 92.08/58.37 & 91.04/60.18 & 91.17/61.21   \\
   \rowcolor{gray!25}  FFT + DSGF    & 86.73/65.30 & 87.03/64.64 & 88.03/63.82 & 90.07/62.66 & 88.96/63.58 & 76.83/71.29   \\
   \cline{2-7}       
                       VAT           & 89.78/64.60 & 82.88/67.25 & 84.30/65.70 & 85.79/64.53 & 86.41/65.18 & 77.40/71.32   \\
   \rowcolor{gray!25}  VAT + DSGF    & 77.79/71.86 & 78.06/70.44 & 79.37/69.21 & 83.15/67.52 & 81.23/69.35 & 77.50/71.86   \\
   \cline{2-7}       
                       VPT           & 79.48/72.54 & 80.89/71.21 & 82.29/69.79 & 84.06/67.93 & 82.59/69.52 & 84.03/69.49   \\
   \rowcolor{gray!25}  VPT + DSGF    & 78.06/72.84 & 77.93/72.83 & 77.92/72.52 & 83.19/69.17 & 82.80/69.54 & 73.41/75.11   \\
\end{tabular}
  \vspace{-3cm}
\end{table*}


\begin{table*}[t]
  \centering
  \caption{Accuracy on Oxford-PETS Dataset. The gray background indicates deploying our method in the current setting.}
  \label{Table19}
    \fontsize{7}{8}\selectfont
  \begin{tabular}{cccccc}
    \multirow{2}{*}{Method} & \multicolumn{5}{c}{ID acc. ($\%$) (ID: Oxford-PETS)} \\
    \cline{2-6}
                    &2-shot   &4-shot   &8-shot  &16-shot  & All-shot  \\
    \hline   
       FFT	         & 49.63 	& 60.32 	& 79.42 	& 87.24 	& 94.25   \\
    \rowcolor{gray!25}    FFT + DSGF	 & {78.74} 	& {84.11} 	& {87.16} 	& {90.68} 	& {94.60}    \\
    \hline   
       VAT	         & 80.01 	& 85.53 	& 87.49 	& 89.18 	& 92.67    \\
    \rowcolor{gray!25}    VAT + DSGF	 & {83.70} 	& {86.89} 	& {88.88} 	& {90.19} 	& {92.94}   \\
    \hline   
       VPT	         & 76.94 	& 85.34 	& 89.86 	& 90.92 	& 93.43    \\
    \rowcolor{gray!25}    VPT + DSGF	 & {81.71} 	& {88.77} 	& {90.30} 	& {91.66} 	& {93.46}   \\
    \hline   
       LPT	         & 80.57 	& 86.32 	& 87.14 	& 88.50 	& 91.44 \\
    \hline

  \end{tabular}
\end{table*}

\end{document}